\documentclass[letterpaper, 10 pt, conference]{ieeeconf}  % Comment this line out if you need a4paper

\IEEEoverridecommandlockouts                              % This command is only needed if 
\makeatletter
\let\NAT@parse\undefined
\makeatother

\usepackage{cite}
\usepackage{amsmath,amssymb,amsfonts}
\usepackage{algorithmic}
\usepackage{graphicx}
\usepackage{pifont}
\usepackage{textcomp}
\usepackage{hyperref}
\usepackage{multirow}
\usepackage{balance}
\usepackage[table,xcdraw]{xcolor}
\usepackage{url}
\usepackage[export]{adjustbox}
\usepackage{comment}
\usepackage{color}
\usepackage{latexsym}
\usepackage{xurl}
\usepackage{mathtools}
\usepackage{caption}
\usepackage{subcaption}

\DeclareMathOperator*{\argmax}{\arg\!\max} % argmax operator

\title{\LARGE \bf
	Multi-Agent Deep Reinforcement Learning for Cooperative and Competitive Autonomous Vehicles using AutoDRIVE Ecosystem
}
\author{Tanmay Samak$^{\star \dagger}$, Chinmay Samak$^{\star \dagger}$ and Venkat Krovi$^{\dagger}$% <-this % stops a space
	\thanks{$^{\star}$These authors contributed equally.}% <-this % stops a space
	\thanks{$^{\dagger}$Automation, Robotics and Mechatronics Lab (ARMLab), Department of Automotive Engineering, Clemson University International Center for Automotive Research (CU-ICAR), Greenville, SC 29607, USA.
		{\tt\small {\{\href{mailto:tsamak@clemson.edu}{tsamak}, \href{mailto:csamak@clemson.edu}{csamak}, \href{mailto:vkrovi@clemson.edu}{vkrovi}\}@clemson.edu}}}%
}

\begin{document}
	
	\maketitle
	\thispagestyle{empty}
	\pagestyle{empty}
	
	%%%%%%%%%%%%%%%%%%%%%%%%%%%%%%%%%%%%%%%%%%%%%%%%%%%%%%%%%%%%%%%%%%%%%%%%%%%%%%%%
	
	\begin{abstract}
		This work presents a modular and parallelizable multi-agent deep reinforcement learning framework for imbibing cooperative as well as competitive behaviors within autonomous vehicles. We introduce AutoDRIVE Ecosystem as an enabler to develop physically accurate and graphically realistic digital twins of Nigel and F1TENTH, two scaled autonomous vehicle platforms with unique qualities and capabilities, and leverage this ecosystem to train and deploy multi-agent reinforcement learning policies. We first investigate an intersection traversal problem using a set of cooperative vehicles (Nigel) that share limited state information with each other in single as well as multi-agent learning settings using a common policy approach. We then investigate an adversarial head-to-head autonomous racing problem using a different set of vehicles (F1TENTH) in a multi-agent learning setting using an individual policy approach. In either set of experiments, a decentralized learning architecture was adopted, which allowed robust training and testing of the approaches in stochastic environments, since the agents were mutually independent and exhibited asynchronous motion behavior. The problems were further aggravated by providing the agents with sparse observation spaces and requiring them to sample control commands that implicitly satisfied the imposed kinodynamic as well as safety constraints. The experimental results for both problem statements are reported in terms of quantitative metrics and qualitative remarks for training as well as deployment phases.\\%
	\end{abstract}
	
	\begin{keywords}
		Multi-Agent Systems, Autonomous Vehicles, Deep Reinforcement Learning, Game Theory, Digital Twins\\%
	\end{keywords}
	
	%%%%%%%%%%%%%%%%%%%%%%%%%%%%%%%%%%%%%%%%%%%%%%%%%%%%%%%%%%%%%%%%%%%%%%%%%%%%%%%%
	
	\section{Introduction}
	\label{Section: Introduction}
	
	In the rapidly evolving landscape of connected and autonomous vehicles (CAVs), the pursuit of intelligent and adaptive driving systems has emerged as a formidable challenge. Multi-Agent Reinforcement Learning (MARL) stands out as a promising avenue in the quest to develop autonomous vehicles capable of navigating complex and dynamic environments, while taking into account the cooperative and/or competitive nature of interactions with their peers. Particularly, cooperative and competitive MARL represent two pivotal approaches to addressing the intricate challenges posed by multi-agent interactions in autonomous driving scenarios. While cooperative MARL encourages agents to collaborate and share information to achieve common objectives, competitive MARL introduces elements of rivalry and adversary among agents, where individual success may come at the expense of others. These paradigms offer crucial insights into the development of autonomous vehicles, and have the potential to reshape the future of transportation.
	
	\begin{figure}[t]
		\centering
		\begin{subfigure}[b]{\linewidth}
			\centering
			\includegraphics[width=\linewidth]{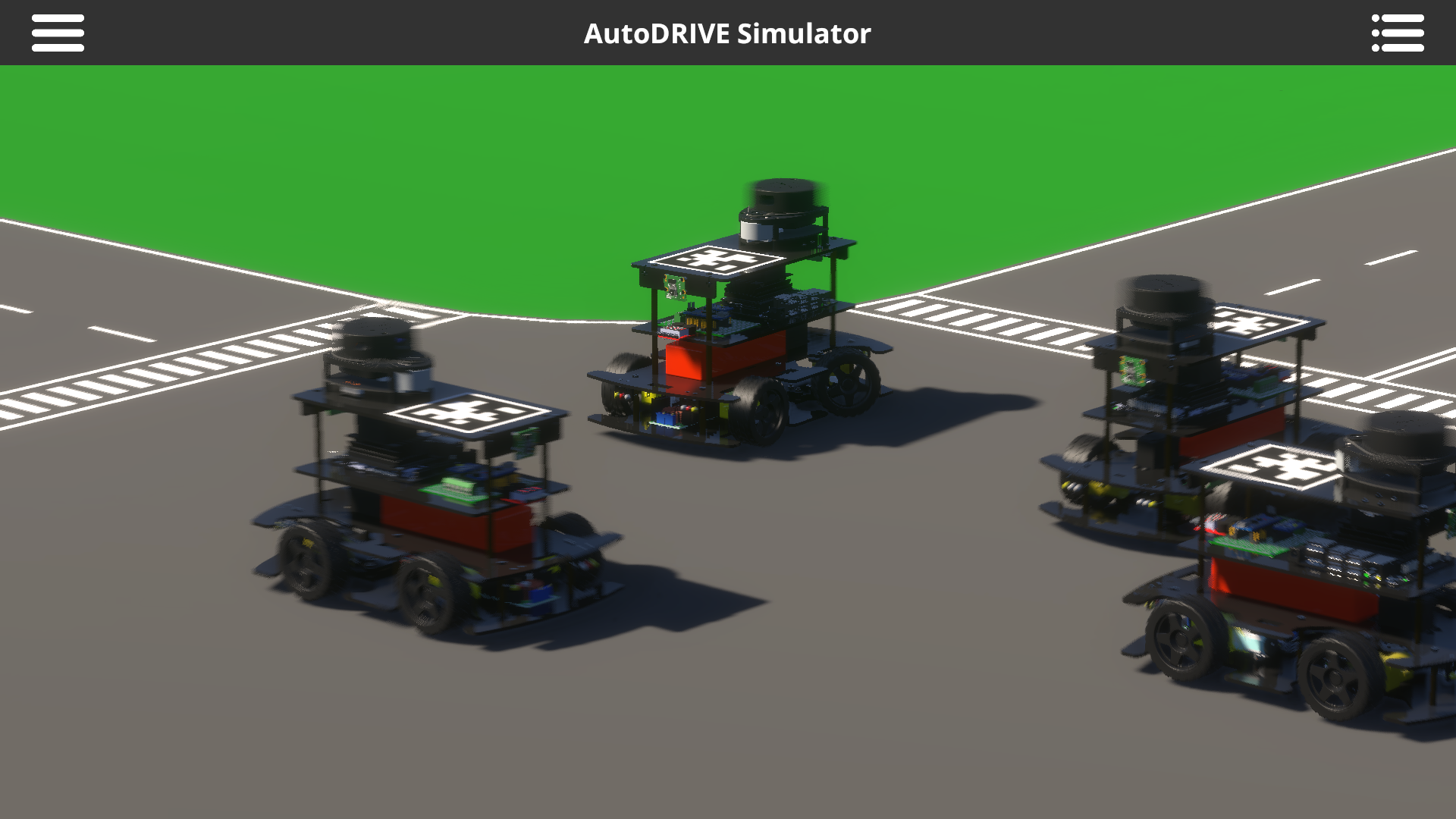}
			\caption{Cooperative MARL using Nigel.}
			\label{fig1a}
		\end{subfigure}
		\hfill
		\begin{subfigure}[b]{\linewidth}
			\centering
			\includegraphics[width=\linewidth]{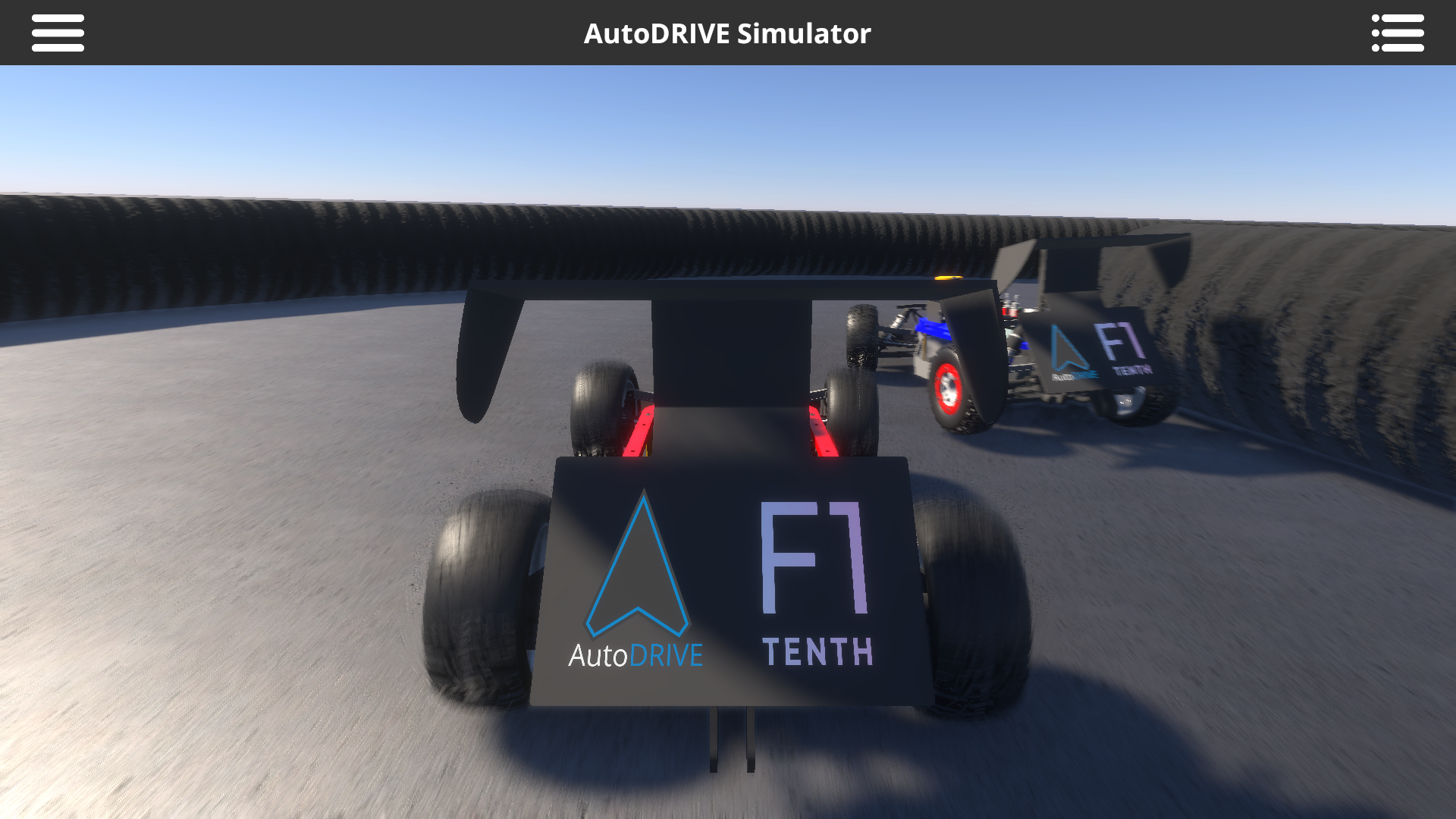}
			\caption{Competitive MARL using F1TENTH.}
			\label{fig1b}
		\end{subfigure}
		\caption{Multi-agent deep reinforcement learning framework using AutoDRIVE Ecosystem.}
		\label{fig1}
	\end{figure}
	
	Cooperative MARL \cite{semnani2020multiagent, long2018optimally, aradi2020survey, wang2020mrcdrl, zhou2019learn, 9316033} fosters an environment in which autonomous vehicles cooperate to accomplish collective objectives such as optimizing traffic flow, enhancing safety, and efficiently navigating road networks. It mirrors real-world situations where vehicles must work together, such as traffic merging, intersection management, or platooning scenarios. Challenges in cooperative MARL include coordinating vehicle actions to minimize congestion, maintaining safety margins, and ensuring smooth interactions between self-interested agents.
	
	\begin{figure*}[t]
		\centering
		\begin{subfigure}[b]{0.245\linewidth}
			\centering
			\includegraphics[width=\linewidth]{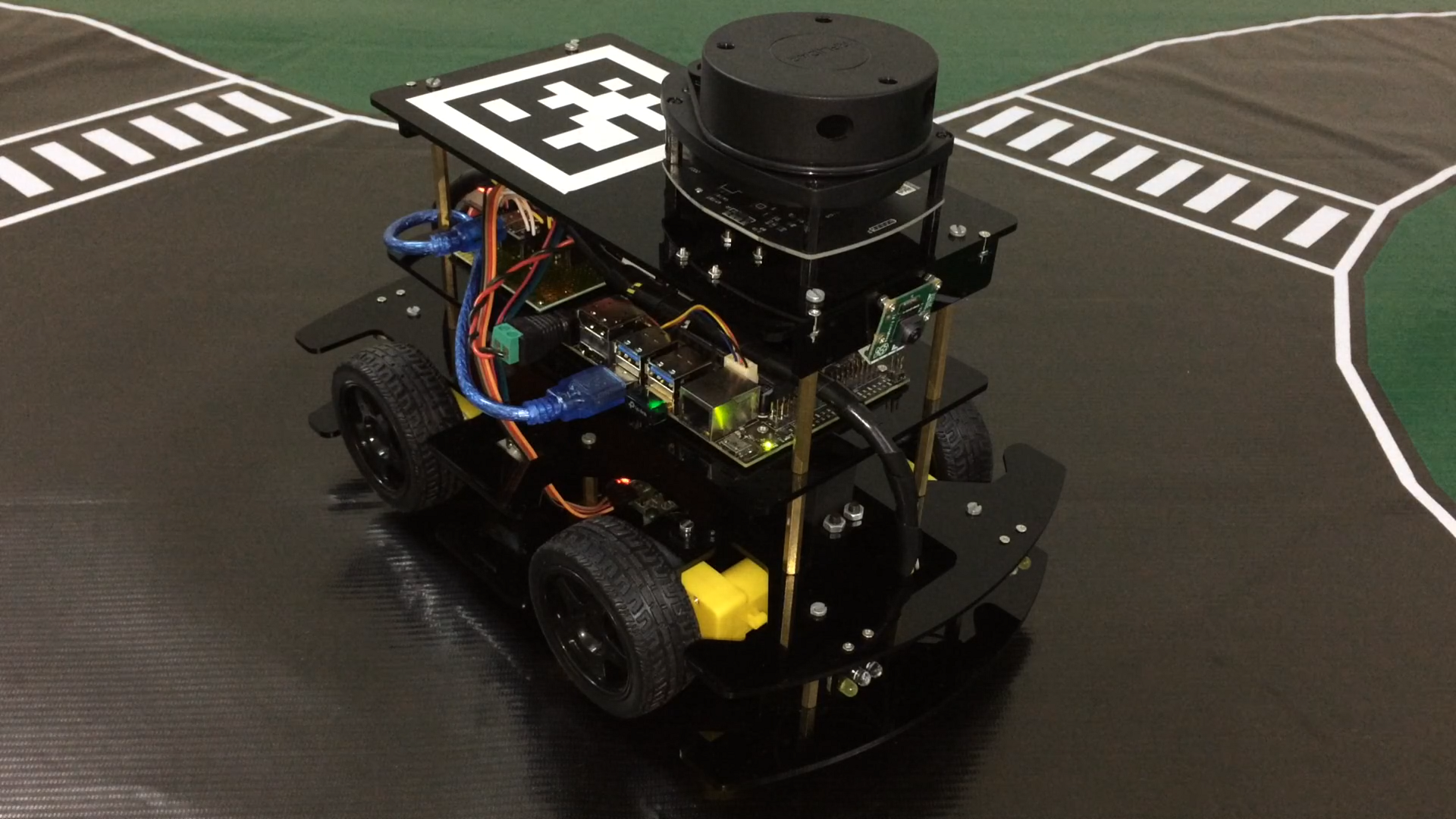}
			\caption{Physical Nigel.}
			\label{fig2a}
		\end{subfigure}
		\hfill
		\begin{subfigure}[b]{0.245\linewidth}
			\centering
			\includegraphics[width=\linewidth]{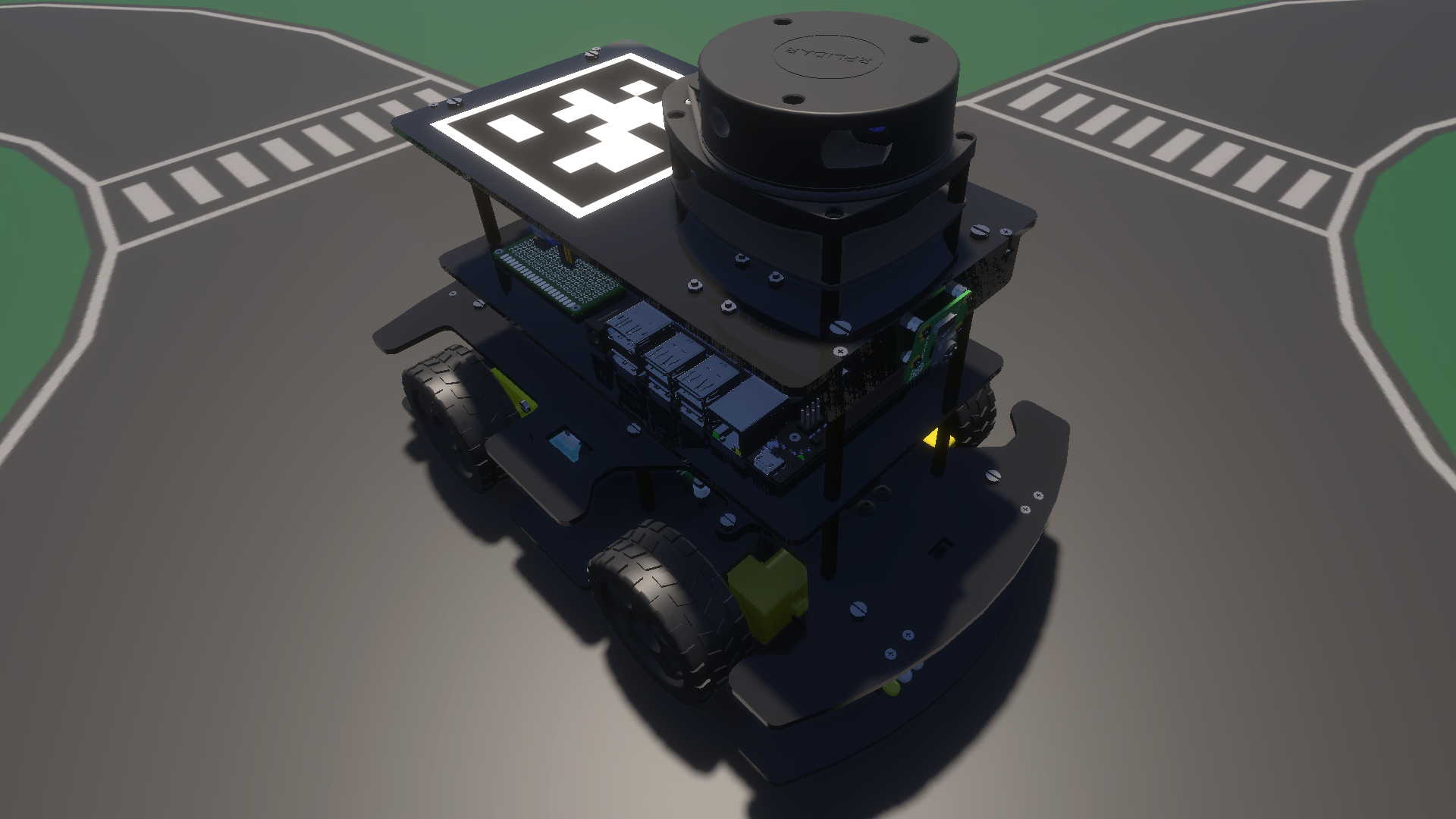}
			\caption{Virtual Nigel.}
			\label{fig2b}
		\end{subfigure}
		\begin{subfigure}[b]{0.245\linewidth}
			\centering
			\includegraphics[width=\linewidth]{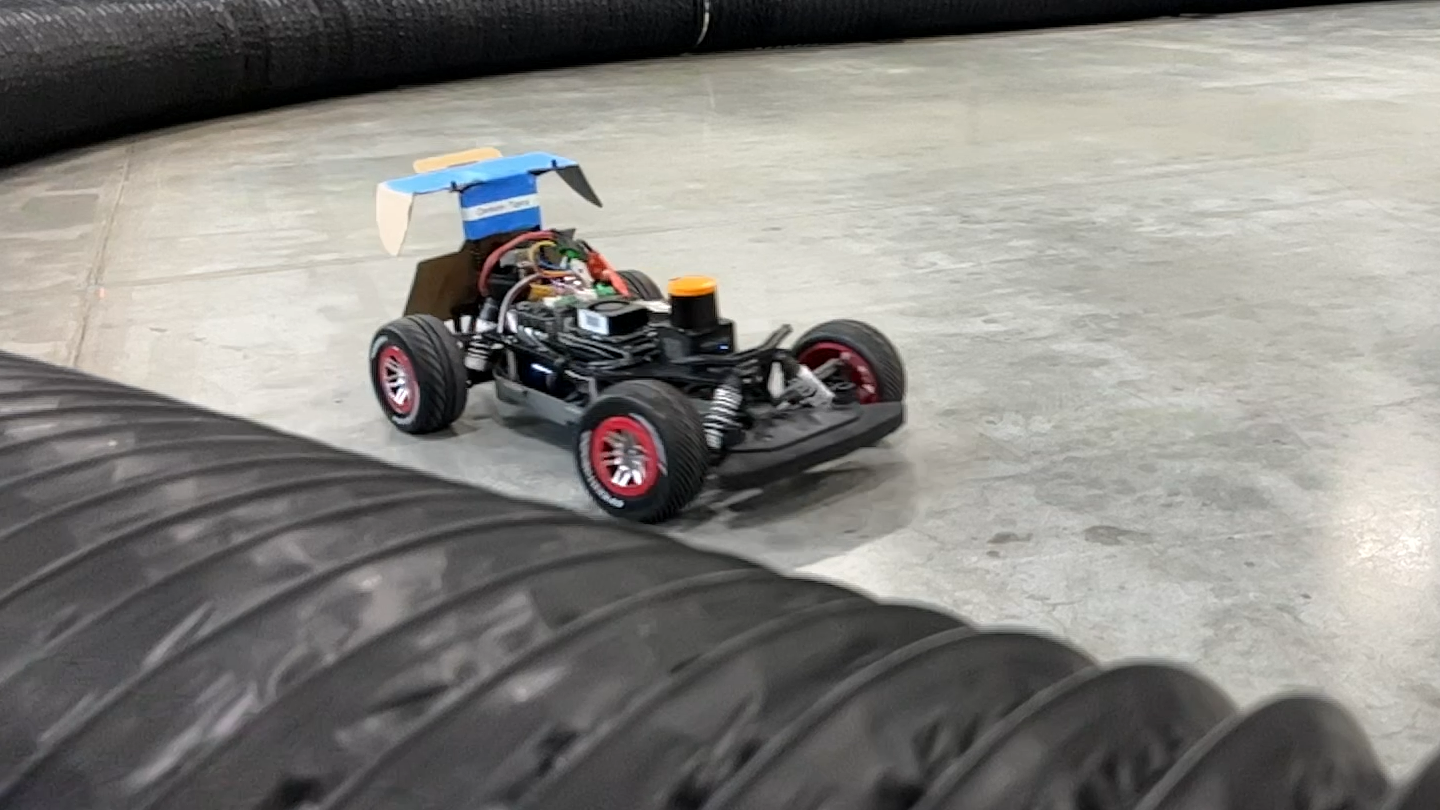}
			\caption{Physical F1TENTH.}
			\label{fig2c}
		\end{subfigure}
		\hfill
		\begin{subfigure}[b]{0.245\linewidth}
			\centering
			\includegraphics[width=\linewidth]{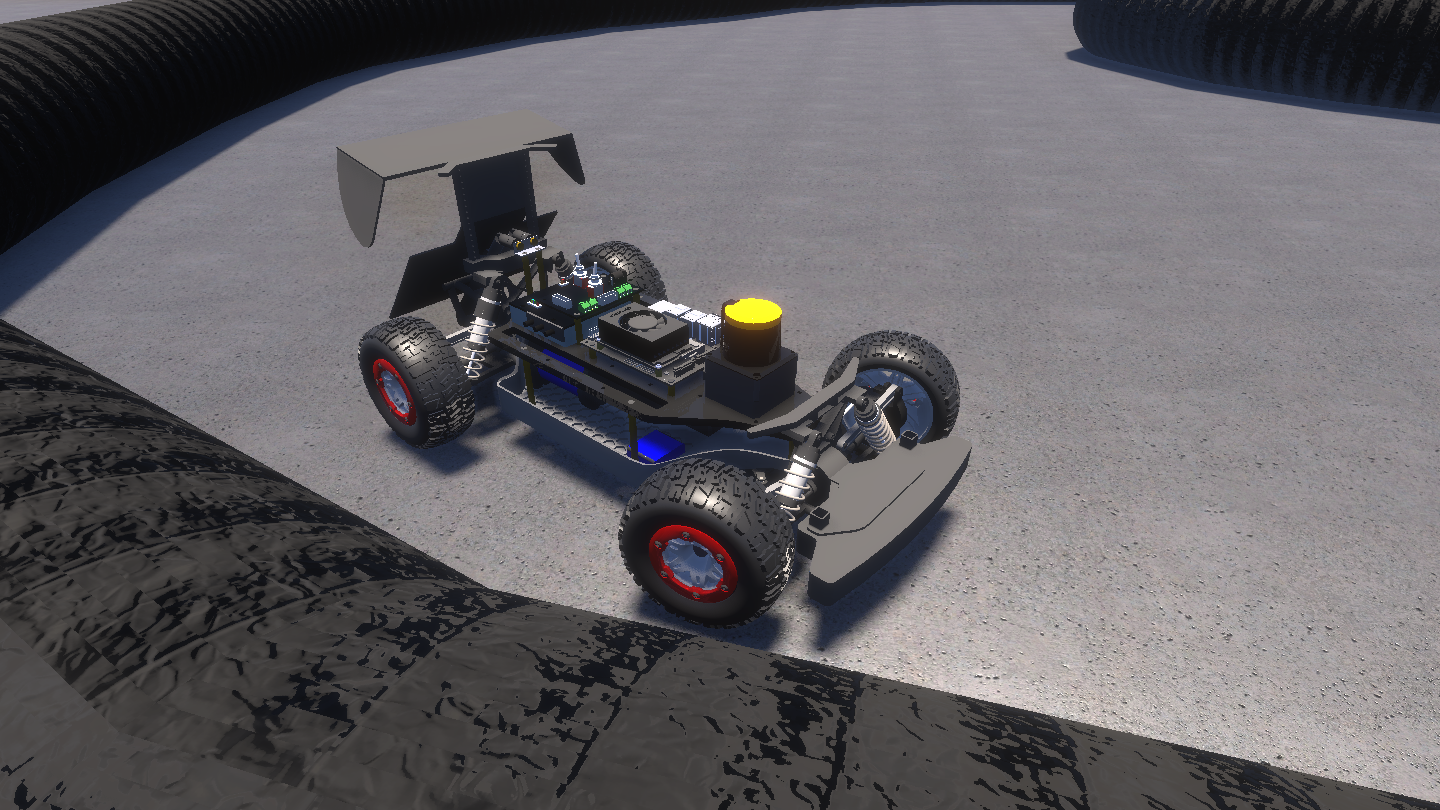}
			\caption{Virtual F1TENTH.}
			\label{fig2d}
		\end{subfigure}
		\caption{Creating the digital twins of Nigel and F1TENTH in AutoDRIVE Simulator.}
		\label{fig2}
	\end{figure*}
	
	On the other hand, competitive MARL \cite{fuchs2020, song2021, AutoRACE-2021, 9790832} introduces a competitive edge to autonomous driving, simulating scenarios such as overtaking, merging in congested traffic, or competitive racing. In this paradigm, autonomous vehicles strive to outperform their counterparts, vying for advantages while navigating complex and dynamic environments. Challenges in competitive MARL encompass strategic decision-making, opponent modeling, and adapting to aggressive driving behaviors while preserving safety standards.
	
	As the field of MARL gains momentum within the realm of autonomous vehicles, it is crucial to comprehensively examine the implications of both cooperative and competitive approaches. In this paper, we present AutoDRIVE Ecosystem \cite{AutoDRIVE2023, AutoDRIVEReport2022} as an enabler to develop physically accurate and graphically realistic digital twins of scaled autonomous vehicles viz. Nigel \cite{10196233} and F1TENTH \cite{F1TENTH2019} in Section \ref{Section: Digital Twin Creation}. We then present the problem formulation, solution approach as well as training and deployment results for a cooperative non-zero-sum use-case of intersection traversal (refer Fig. \hyperref[fig1]{\ref*{fig1}(a)}) in Section \ref{Section: Cooperative Multi-Agent Scenario} and a competitive zero-sum use-case of head-to-head autonomous racing (refer Fig. \hyperref[fig1]{\ref*{fig1}(b)}) in Section \ref{Section: Competitive Multi-Agent Scenario}. Finally we present an overall summary of our work with some concluding remarks on either case-studies.
	
	% %%%%%%%%%%%%%%%%%%%%%%%%%%%%%%%%%%%%%%%%%%%%%%%%%%%%%%%%%%%%%%%%%%%%%%%%%%%%%%%%
	
	\section{Digital Twin Creation}
	\label{Section: Digital Twin Creation}
	
	We leveraged AutoDRIVE Simulator \cite{AutoDRIVESimulator2021, AutoDRIVESimulatorReport2020} to develop digital twin models of Nigel as well as F1TENTH. It is to be noted, however, that this work utilizes the said models in the capacity of virtual prototyping, but we seek to further investigate emerging possibilities of utilizing the digital thread for harnessing the true potential of digital twins.
	
	\subsection{Vehicle Dynamics Models}
	\label{Sub-Section: Vehicle Dynamics Models}
	
	The vehicle model is a combination of a rigid body and a collection of sprung masses $^iM$, where the total mass of the rigid body is defined as $M=\sum{^iM}$. The rigid body's center of mass, $X_{COM} = \frac{\sum{{^iM}*{^iX}}}{\sum{^iM}}$, connects these representations, with $^iX$ representing the coordinates of the sprung masses.
	
	The suspension force acting on each sprung mass is computed as ${^iM} * {^i{\ddot{Z}}} + {^iB} * ({^i{\dot{Z}}}-{^i{\dot{z}}}) + {^iK} * ({^i{Z}}-{^i{z}})$, where $^iZ$ and $^iz$ are the displacements of sprung and unsprung masses, and $^iB$ and $^iK$ are the damping and spring coefficients of the $i$-th suspension, respectively.
	
	The vehicle's wheels are also treated as rigid bodies with mass $m$, subject to gravitational and suspension forces: ${^im} * {^i{\ddot{z}}} + {^iB} * ({^i{\dot{z}}}-{^i{\dot{Z}}}) + {^iK} * ({^i{z}}-{^i{Z}})$.
	
	Tire forces are computed based on the friction curve for each tire, represented as $\left\{\begin{matrix} {^iF_{t_x}} = F(^iS_x) \\{^iF_{t_y}} = F(^iS_y) \\ \end{matrix}\right.$, where $^iS_x$ and $^iS_y$ are the longitudinal and lateral slips of the $i$-th tire, respectively. The friction curve is approximated using a two-piece cubic spline, defined as $F(S) = \left\{\begin{matrix} f_0(S); \;\; S_0 \leq S < S_e \\ f_1(S); \;\; S_e \leq S < S_a \\ \end{matrix}\right.$, where $f_k(S) = a_k*S^3+b_k*S^2+c_k*S+d_k$ is a cubic polynomial function. The first segment of the spline ranges from zero $(S_0,F_0)$ to an extremum point $(S_e,F_e)$, while the second segment ranges from the extremum point $(S_e, F_e)$ to an asymptote point $(S_a, F_a)$.
	
	The tire slip is influenced by factors including tire stiffness $^iC_\alpha$, steering angle $\delta$, wheel speeds $^i\omega$, suspension forces $^iF_s$, and rigid-body momentum $^iP$. These factors impact the longitudinal and lateral components of the vehicle's linear velocity. The longitudinal slip $^iS_x$ of the $i$-th tire is calculated by comparing the longitudinal components of the surface velocity of the $i$-th wheel (i.e., longitudinal linear velocity of the vehicle) $v_x$ with the angular velocity $^i\omega$ of the $i$-th wheel: ${^iS_x} = \frac{{^ir}*{^i\omega}-v_x}{v_x}$. The lateral slip $^iS_y$ depends on the tire's slip angle $\alpha$ and is determined by comparing the longitudinal $v_x$ (forward velocity) and lateral $v_y$ (side-slip velocity) components of the vehicle's linear velocity: ${^iS_y} = \tan(\alpha) = \frac{v_y}{\left| v_x \right|}$.
	
	\subsection{Sensor Models}
	\label{Sub-Section: Sensor Models}
	
	The simulated vehicles can be equipped with the physically accurate interoceptive as well as exteroceptive sensing modalities. Specifically, the throttle ($\tau$) and steering ($\delta$) sensors are simulated using a straightforward feedback loop.
	
	Incremental encoders are simulated by measuring the rotation of the rear wheels (i.e., the output shaft of driving actuators): $^iN_{ticks} = {^iPPR} * {^iGR} * {^iN_{rev}}$, where $^iN_{ticks}$ represents the ticks measured by the $i$-th encoder, $^iPPR$ is the base resolution (pulses per revolution) of the $i$-th encoder, $^iGR$ is the gear ratio of the $i$-th motor, and $^iN_{rev}$ represents the number of revolutions of the output shaft of the $i$-th motor.
	
	The Inertial Positioning System (IPS) and Inertial Measurement Unit (IMU) are simulated based on temporally-coherent rigid-body transform updates of the vehicle $\{v\}$ with respect to the world $\{w\}$: ${^w\mathbf{T}_v} = \left[\begin{array}{c | c} \mathbf{R}_{3 \times 3} & \mathbf{t}_{3 \times 1} \\ \hline \mathbf{0}_{1 \times 3} & 1 \end{array}\right] \in SE(3)$. The IPS provides 3-DOF positional coordinates $\{x,y,z\}$ of the vehicle, while the IMU supplies linear accelerations $\{a_x,a_y,a_z\}$, angular velocities $\{\omega_x,\omega_y,\omega_z\}$, and 3-DOF orientation data for the vehicle, either as Euler angles $\{\phi_x,\theta_y,\psi_z\}$ or as a quaternion $\{q_0,q_1,q_2,q_3\}$.
	
	The LIDAR simulation employs iterative ray-casting \texttt{raycast}\{$^w\mathbf{T}_l$, $\vec{\mathbf{R}}$, $r_{max}$\} for each angle $\theta \in \left [ \theta_{min}:\theta_{res}:\theta_{max} \right ]$ at an approximate update rate of 7 Hz. Here, ${^w\mathbf{T}_l} = {^w\mathbf{T}_v} * {^v\mathbf{T}_l} \in SE(3)$ represents the relative transformation of the LIDAR \{$l$\} with respect to the vehicle \{$v$\} and the world \{$w$\}, $\vec{\mathbf{R}} = \left [r_{max}*sin(\theta) \;\; r_{min}*cos(\theta) \;\; 0 \right ]^T$ defines the direction vector of each ray-cast $R$, where $r_{min}=$ 0.15 m and $r_{max}=$ 12 m denote the minimum and maximum linear ranges of the LIDAR, $\theta_{min}=0^\circ$ and $\theta_{max}=360^\circ$ set the minimum and maximum angular ranges of the LIDAR, and $\theta_{res}=1^\circ$ represents the angular resolution of the LIDAR. The laser scan ranges are determined by checking ray-cast hits and then applying a threshold to the minimum linear range of the LIDAR, calculated as \texttt{ranges[i]}$=\begin{cases} \texttt{hit.dist} & \text{ if } \texttt{ray[i].hit} \text{ and } \texttt{hit.dist} \geq r_{min} \\ \infty & \text{ otherwise} \end{cases}$, where \texttt{ray.hit} is a Boolean flag indicating whether a ray-cast hits any colliders in the scene, and \texttt{hit.dist}$=\sqrt{(x_{hit}-x_{ray})^2 + (y_{hit}-y_{ray})^2 + (z_{hit}-z_{ray})^2}$ calculates the Euclidean distance from the ray-cast source $\{x_{ray}, y_{ray}, z_{ray}\}$ to the hit point $\{x_{hit}, y_{hit}, z_{hit}\}$.
	
	The simulated physical cameras are parameterized by their focal length ($f=$ 3.04 mm), sensor size ($\{s_x, s_y\} = $ \{3.68, 2.76\} mm), target resolution (default = 720p), as well as the distances to the near and far clipping planes ($N=$ 0.01 m and $F=$ 1000 m). The viewport rendering pipeline for the simulated cameras operates in three stages. First, the camera view matrix $\mathbf{V} \in SE(3)$ is computed by obtaining the relative homogeneous transform of the camera $\{c\}$ with respect to the world $\{w\}$: $\mathbf{V} = \begin{bmatrix} r_{00} & r_{01} & r_{02} & t_{0} \\ r_{10} & r_{11} & r_{12} & t_{1} \\ r_{20} & r_{21} & r_{22} & t_{2} \\ 0 & 0 & 0 & 1 \\ \end{bmatrix}$, where $r_{ij}$ and $t_i$ denote the rotational and translational components, respectively. Next, the camera projection matrix $\mathbf{P} \in \mathbb{R}^{4 \times 4}$ is calculated to project world coordinates into image space coordinates: $\mathbf{P} = \begin{bmatrix} \frac{2*N}{R-L} & 0 & \frac{R+L}{R-L} & 0 \\ 0 & \frac{2*N}{T-B} & \frac{T+B}{T-B} & 0 \\ 0 & 0 & -\frac{F+N}{F-N} & -\frac{2*F*N}{F-N} \\ 0 & 0 & -1 & 0 \\ \end{bmatrix}$, where $N$ and $F$ represent the distances to the near and far clipping planes of the camera, and $L$, $R$, $T$, and $B$ denote the left, right, top, and bottom offsets of the sensor. The camera parameters $\{f,s_x,s_y\}$ are related to the terms of the projection matrix as follows: $f = \frac{2*N}{R-L}$, $a = \frac{s_y}{s_x}$, and $\frac{f}{a} = \frac{2*N}{T-B}$. The perspective projection from the simulated camera's viewport is given as $\mathbf{C} = \mathbf{P}*\mathbf{V}*\mathbf{W}$, where $\mathbf{C} = \left [x_c\;\;y_c\;\;z_c\;\;w_c \right ]^T$ represents image space coordinates, and $\mathbf{W} = \left [x_w\;\;y_w\;\;z_w\;\;w_w \right ]^T$ represents world coordinates. Finally, this camera projection is transformed into normalized device coordinates (NDC) by performing perspective division (i.e., dividing throughout by $w_c$), leading to a viewport projection achieved by scaling and shifting the result and then utilizing the rasterization process of the graphics API (e.g., DirectX for Windows, Metal for macOS, and Vulkan for Linux). Additionally, a post-processing step simulates lens and film effects, such as lens distortion, depth of field, exposure, ambient occlusion, contact shadows, bloom, motion blur, film grain, chromatic aberration, etc.
	
	\subsection{Actuator Models}
	\label{Sub-Section: Actuator Models}
	
	The vehicle's motion is controlled by driving and steering actuators, with response delays and saturation limits matched to their real-world counterparts by tuning their torque profiles and actuation limits.
	
	The driving actuators propel the rear/front/all wheels by applying a torque, calculated as ${^i\tau_{drive}} = {^iI_w}*{^i\dot{\omega}_w}$, where ${^iI_w} = \frac{1}{2}*{^im_w}*{^i{r_w}^2}$ represents the moment of inertia, $^i\dot{\omega}_w$ is the angular acceleration, $^im_w$ is the mass, and $^ir_w$ is the radius of the $i$-th wheel. Additionally, the driving actuators simulate holding torque by applying an idle motor torque equivalent to the braking torque, i.e., ${^i\tau_{idle}} = {^i\tau_{brake}}$.
	
	The front wheels are steered using a steering actuator that generates a torque proportional to the required angular acceleration, given by $\tau_{steer} = I_{steer}*\dot{\omega}_{steer}$. The individual turning angles, $\delta_l$ and $\delta_r$, for the left and right wheels, respectively, are computed based on the commanded steering angle $\delta$, utilizing the Ackermann steering geometry defined by the wheelbase $l$ and track width $w$, as follows: $\left\{\begin{matrix} \delta_l = \textup{tan}^{-1}\left(\frac{2*l*\textup{tan}(\delta)}{2*l+w*\textup{tan}(\delta)}\right) \\ \delta_r = \textup{tan}^{-1}\left(\frac{2*l*\textup{tan}(\delta)}{2*l-w*\textup{tan}(\delta)}\right) \end{matrix}\right.$
	
	\subsection{Environment Models}
	\label{Sub-Section: Environment Models}
	
	At each time step, the simulator conducts mesh-mesh interference detection and computes contact forces, frictional forces, momentum transfer, as well as linear and angular drag acting on all rigid bodies. Simulated environments can be established using one of the following approaches:
	\begin{itemize}
		\item \textit{AutoDRIVE IDK:} Custom scenarios and maps can be crafted by utilizing the modular and adaptable Infrastructure Development Kit (IDK). This kit provides the flexibility to configure terrain modules, road networks, obstruction modules, and traffic elements. Specifically, the intersection traversal scenario was developed using AutoDRIVE IDK.
		
		\item \textit{Plug-In Scenarios:} AutoDRIVE Simulator supports third-party tools, such as RoadRunner \cite{RoadRunner2021}, and open standards like OpenSCENARIO \cite{OpenSCENARIO2021} and OpenDRIVE \cite{OpenDRIVE2021}). This allows users to incorporate a diverse range of plugins, packages, and assets in several standard formats for creating or customizing driving scenarios. Particularly, the autonomous racing scenario was created based on the binary occupancy grid map of a real-world F1TENTH racetrack called ``Proto'' using a third-party 3D modelling software, which was then imported into AutoDRIVE Simulator and post-processed with physical as well as graphical enhancements to make it ``sim-ready''.
		
		\item \textit{Unity Terrain Integration:} Since the AutoDRIVE Simulator is built atop the Unity \cite{Unity2021} game engine, it seamlessly supports scenario design and development through Unity Terrain \cite{UnityTerrain2021}. Users have the option to define terrain meshes, textures, heightmaps, vegetation, skyboxes, wind effects, and more, allowing design of both on-road and off-road scenarios. This option is well-suited for modelling full-scale environments.
	\end{itemize}
	
	% %%%%%%%%%%%%%%%%%%%%%%%%%%%%%%%%%%%%%%%%%%%%%%%%%%%%%%%%%%%%%%%%%%%%%%%%%%%%%%%%
	
	\section{Cooperative Multi-Agent Scenario}
	\label{Section: Cooperative Multi-Agent Scenario}
	
	Inspired by \cite{9316033}, this use-case encompassed both single-agent and multi-agent learning scenarios, where each agent's objective was autonomous traversal of a 4-lane, 4-way intersection without collisions or lane boundary violations. Each vehicle possessed intrinsic state information and received limited state information about its peers; no external sensing modalities were employed. A deep neural network policy was independently trained for each scenario, guiding the agents through the intersection safely. The entire case-study was developed using an integrated ML framework \cite{MLAgents2018} within AutoDRIVE Simulator.
	
	\subsection{Problem Formulation}
	\label{Sub-Section: Problem Formulation I}
	
	In \textit{single-agent learning scenario}, only the ego vehicle learned to traverse the intersection, while peer vehicles were controlled at different velocities using a simple heuristic. Peer vehicles shared their states with the ego vehicle via V2V communication. All vehicles were reset together, making this scenario quite deterministic.
	
	In \textit{multi-agent learning scenario}, all vehicles learned to traverse the intersection simultaneously in a decentralized manner. Vehicles shared their states with each other via V2V communication and were reset independently, resulting in a highly stochastic scenario.
	
	In both the scenarios, the challenge revolved around autonomous navigation in an unknown environment. The exact structure/map of the environment was not known to any agent. Consequently, this decision-making problem was framed as a Partially Observable Markov Decision Process (POMDP), which captured hidden state information through environmental observations.
	
	\subsection{State Space}
	\label{Sub-Section: State Space I}
	
	As previously discussed, the state space $S$ for intersection traversal problem could be divided into observable $s^o \subset S$ and hidden $s^h \subset S$ components. The observable component included the vehicle's 2D pose and velocity, denoted as $s_{t}^{o}=\left [ p_{x}, p_{y}, \psi, v \right ]{t} \in \mathbb{R}^{4}$. The hidden component encompassed the agent's goal coordinates, represented as $s{t}^{h}=\left [ g_{x}, g_{y} \right ]_{t} \in \mathbb{R}^{2}$. Thus, each agent could observe the pose and velocity of its peers (via V2V communication) but kept its own goal location hidden from others. Consequently, the complete state space of an agent participating in this problem was a vector containing all observable and hidden states:
	
	\begin{equation}
		\label{Equation: 5.31}
		s_{t} = \left [ s_{t}^{o}, s_{t}^{h} \right ]
	\end{equation}
	
	\subsection{Observation Space}
	\label{Sub-Section: Observation Space I}
	
	Based on the state space defined in Equation \ref{Equation: 5.31}, each agent employed an appropriate subset of its sensor suite to collect observations (as per Equation \ref{Equation: 5.32}). This included Inertial Positioning System (IPS) for positional coordinates $\left [ p_{x}, p_{y} \right ]{t} \in \mathbb{R}^{2}$, Inertial Measurement Unit (IMU) for yaw $\psi{t} \in \mathbb{R}^{1}$, and incremental encoders for estimating vehicle velocity $v_{t} \in \mathbb{R}^{1}$. Each agent $i$ (where $0<i<N$) was provided with an observation vector of the form:
	
	\begin{equation}
		\label{Equation: 5.32}
		o_{t}^{i} = \left [ g^{i}, \tilde{p}^{i}, \tilde{\psi}^{i}, \tilde{v}^{i} \right ]_{t} \in \mathbb{R}^{2+4(N-1)}
	\end{equation}
	
	This formulation allowed $g_{t}^{i} = \left [ g_{x}^{i}-p_{x}^{i}, g_{y}^{i}-p_{y}^{i} \right ]{t} \in \mathbb{R}^{2}$ to represent the ego agent's goal location relative to itself, $\tilde{p}{t}^{i} = \left [ p_{x}^{j}-p_{x}^{i}, p_{y}^{j}-p_{y}^{i} \right ]{t} \in \mathbb{R}^{2(N-1)}$ to denote the position of every peer agent relative to the ego agent, $\tilde{\psi}{t}^{i} = \psi_{t}^{j}-\psi_{t}^{i} \in \mathbb{R}^{N-1}$ to express the yaw of every peer agent relative to the ego agent, and $\tilde{v}{t}^{i} = v{t}^{j} \in \mathbb{R}^{N-1}$ to indicate the velocity of every peer agent. Here, $i$ represented the ego agent, and $j \in \left [ 0, N-1 \right ]$ represented every other (peer) agent in the scene, with a total of $N$ agents. 
	
	\subsection{Action Space}
	\label{Sub-Section: Action Space I}
	
	The vehicles were designed as non-holonomic rear-wheel-drive models featuring an Ackermann steering mechanism. As a result, the complete action space of an agent comprised longitudinal (throttle/brake) and lateral (steering) motion control commands. For longitudinal control, the throttle command $\tau_t$ was set to 80\% of its upper saturation limit. The steering command $\delta_t$ was discretized as $\delta_t \in \left \{ -1, 0, 1 \right \}$ and was the sole active control source for safely navigating the intersection, as expressed in Equation \ref{Equation: 5.33}:
	
	\begin{equation}
		\label{Equation: 5.33}
		a_t = \delta_t \in \mathbb{R}^{1}
	\end{equation}
	
	\subsection{Reward Function}
	\label{Sub-Section: Reward Function I}
	
	The extrinsic reward function (as shown in Equation \ref{Equation: 5.34}) was designed to reward each agent with $r_{goal}=+1$ for successfully traversing the intersection. Alternatively, it penalized agents proportionally to their distance from the goal, represented as $k_p * \left \| g_{t}^{i} \right \|_{2}$, for collisions or lane boundary violations. The penalty constant $k_p$ was set to 0.425, resulting in a maximum penalty of 1.
	
	\begin{align}
		\label{Equation: 5.34}
		r_{t}^{i} = \begin{cases}
			r_{goal}; \textup{\qquad\qquad if traversed the intersection safely}\\
			-k_p * \left \| g_{t}^{i} \right \|_{2}; \textup{\;\, if collided or overstepped lanes}
		\end{cases}
	\end{align}
	
	This encouraged agents to get closer to their respective goals, reducing penalties and ultimately leading to a positive reward, $r{goal}$. This approach not only expedited convergence but also restricted reward hacking.
	
	\subsection{Optimization Problem}
	\label{Sub-Section: Optimization Problem I}
	
	The task of intersection traversal, with collision avoidance and lane-keeping constraints, was addressed through the extrinsic reward function described in Equation \ref{Equation: 5.34}. This function motivated each individual agent to maximize the expected future discounted reward (as per Equation \ref{Equation: 5.35}) by learning a policy $\pi_\theta \left(a_t|o_t\right)$. Over time, this policy transitioned into the optimal policy $\pi^*$.
	
	\begin{align}
		\label{Equation: 5.35}
		\argmax_{\pi_\theta \left(a_t|o_t\right)} \quad &\mathbb{E}\left [ \sum_{t=0}^{\infty} \gamma^t r_t \right ]
	\end{align}
	
	\begin{figure*}[t]
		\centering
		\begin{subfigure}[b]{0.16\linewidth}
			\centering
			\includegraphics[width=\linewidth]{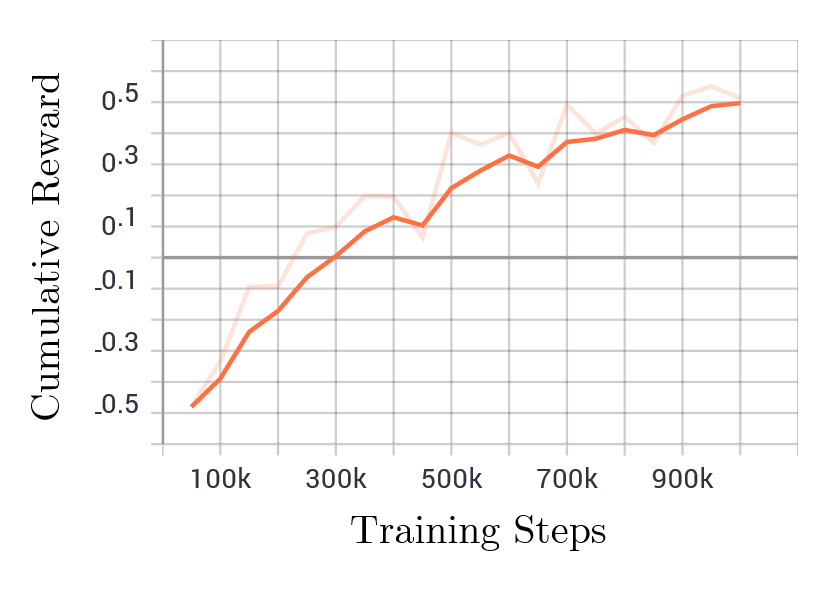}
			\caption{}
			\label{fig3a}
		\end{subfigure}
		\hfill
		\begin{subfigure}[b]{0.16\linewidth}
			\centering
			\includegraphics[width=\linewidth]{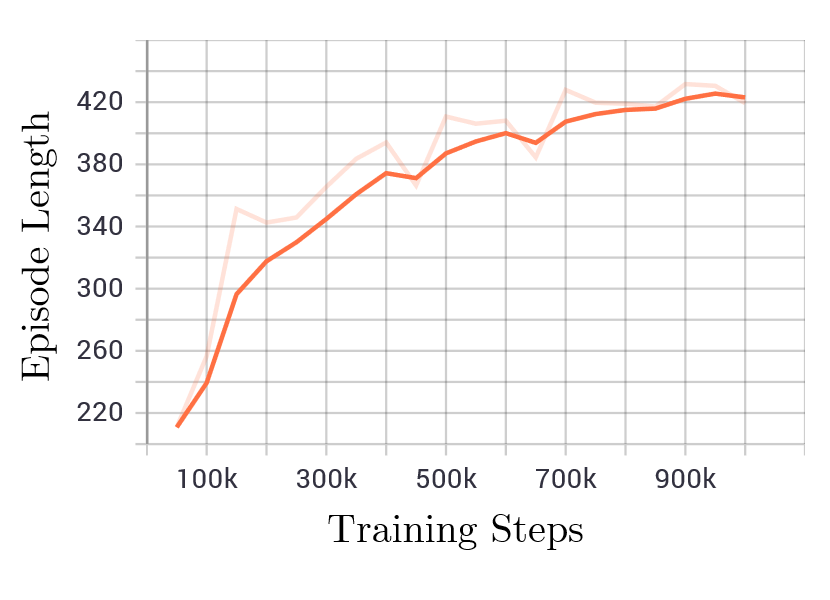}
			\caption{}
			\label{fig3b}
		\end{subfigure}
		\hfill
		\begin{subfigure}[b]{0.16\linewidth}
			\centering
			\includegraphics[width=\linewidth]{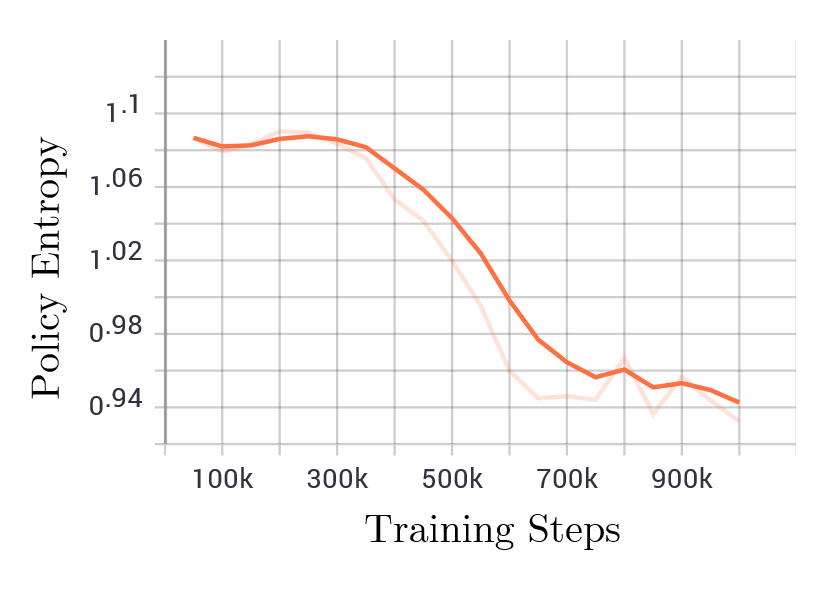}
			\caption{}
			\label{fig3c}
		\end{subfigure}
		\hfill
		\begin{subfigure}[b]{0.16\linewidth}
			\centering
			\includegraphics[width=\linewidth]{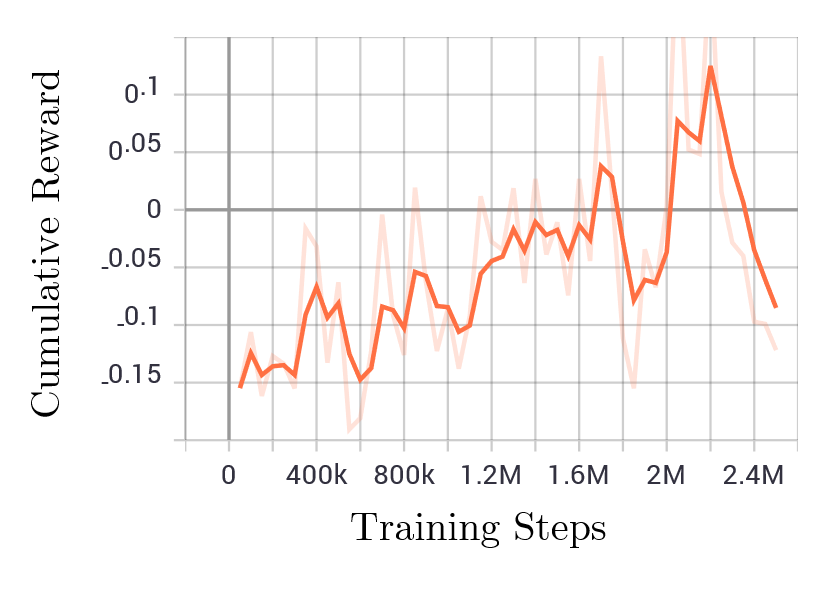}
			\caption{}
			\label{fig3d}
		\end{subfigure}
		\hfill
		\begin{subfigure}[b]{0.16\linewidth}
			\centering
			\includegraphics[width=\linewidth]{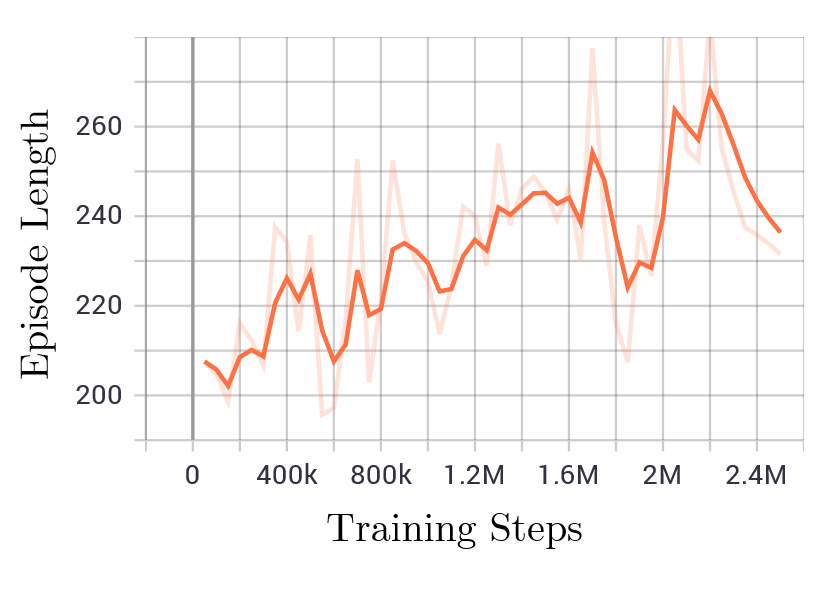}
			\caption{}
			\label{fig3e}
		\end{subfigure}
		\hfill
		\begin{subfigure}[b]{0.16\linewidth}
			\centering
			\includegraphics[width=\linewidth]{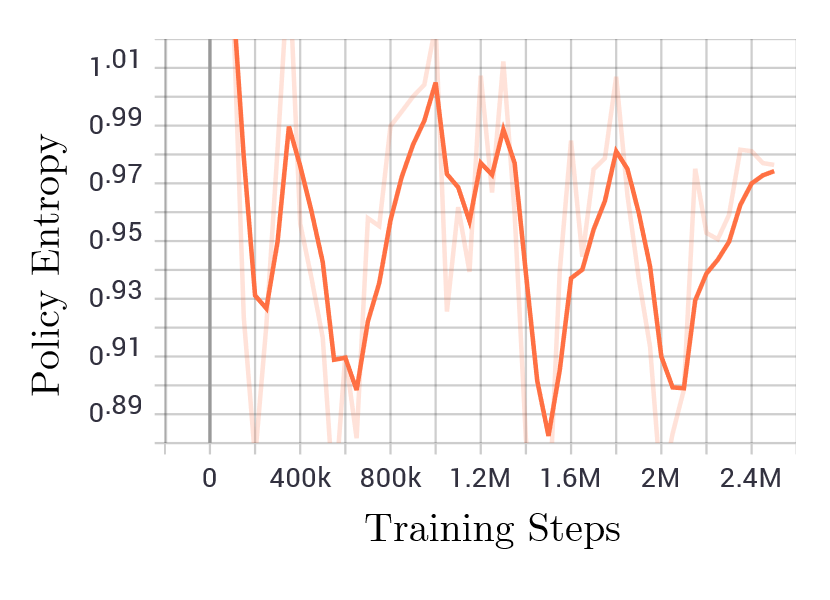}
			\caption{}
			\label{fig3f}
		\end{subfigure}
		\caption{Training results for (a)-(c) single-agent and (d)-(f) multi-agent intersection traversal scenarios: (a) and (d) denote cumulative reward, (b) and (e) denote episode length, while (c) and (f) denote policy entropy w.r.t. training steps.}
		\label{fig3}
	\end{figure*}
	
	\begin{figure*}[t]
		\centering
		\begin{subfigure}[b]{0.16\linewidth}
			\centering
			\includegraphics[width=\linewidth]{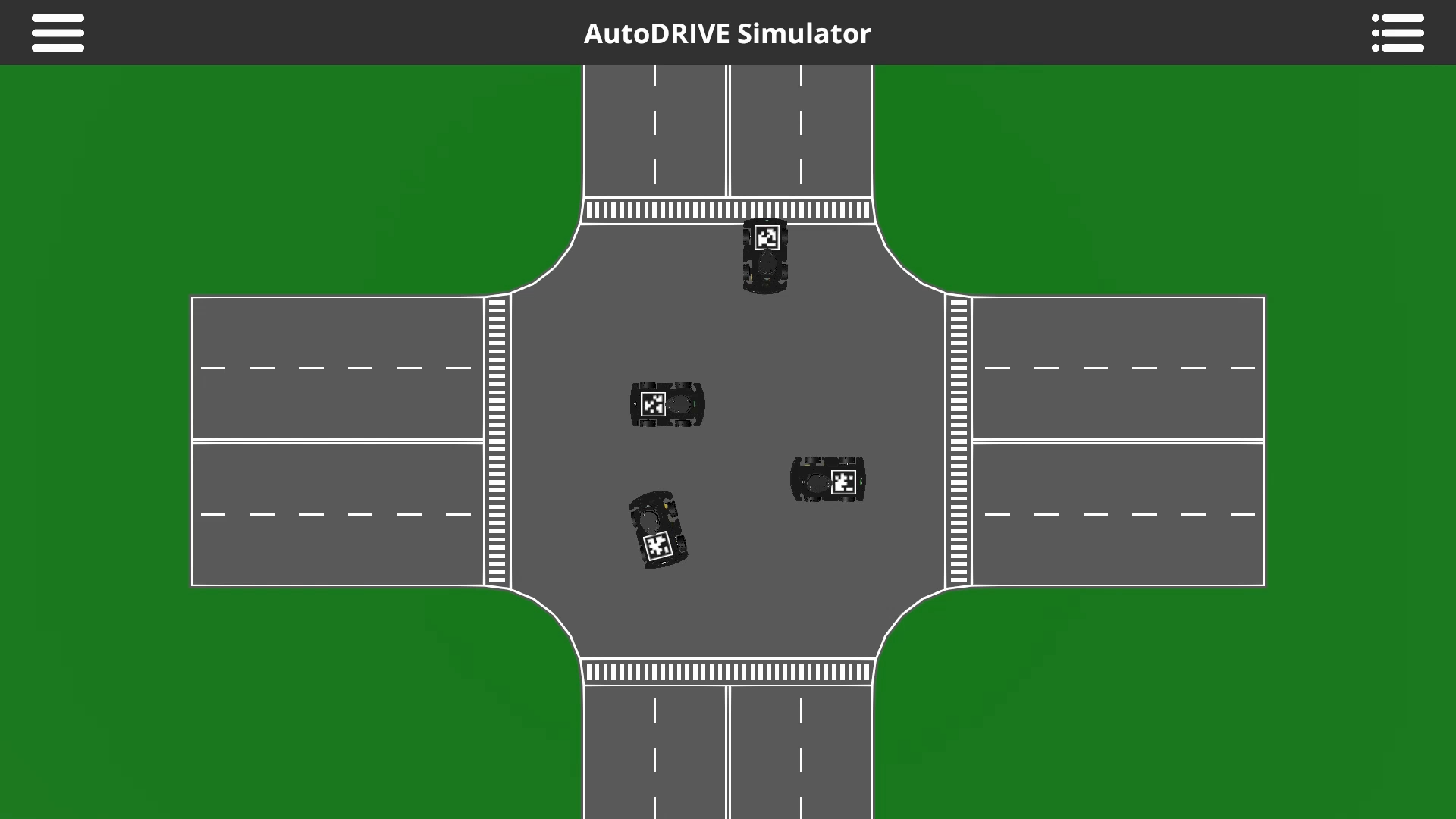}
			\caption{}
			\label{fig4a}
		\end{subfigure}
		\hfill
		\begin{subfigure}[b]{0.16\linewidth}
			\centering
			\includegraphics[width=\linewidth]{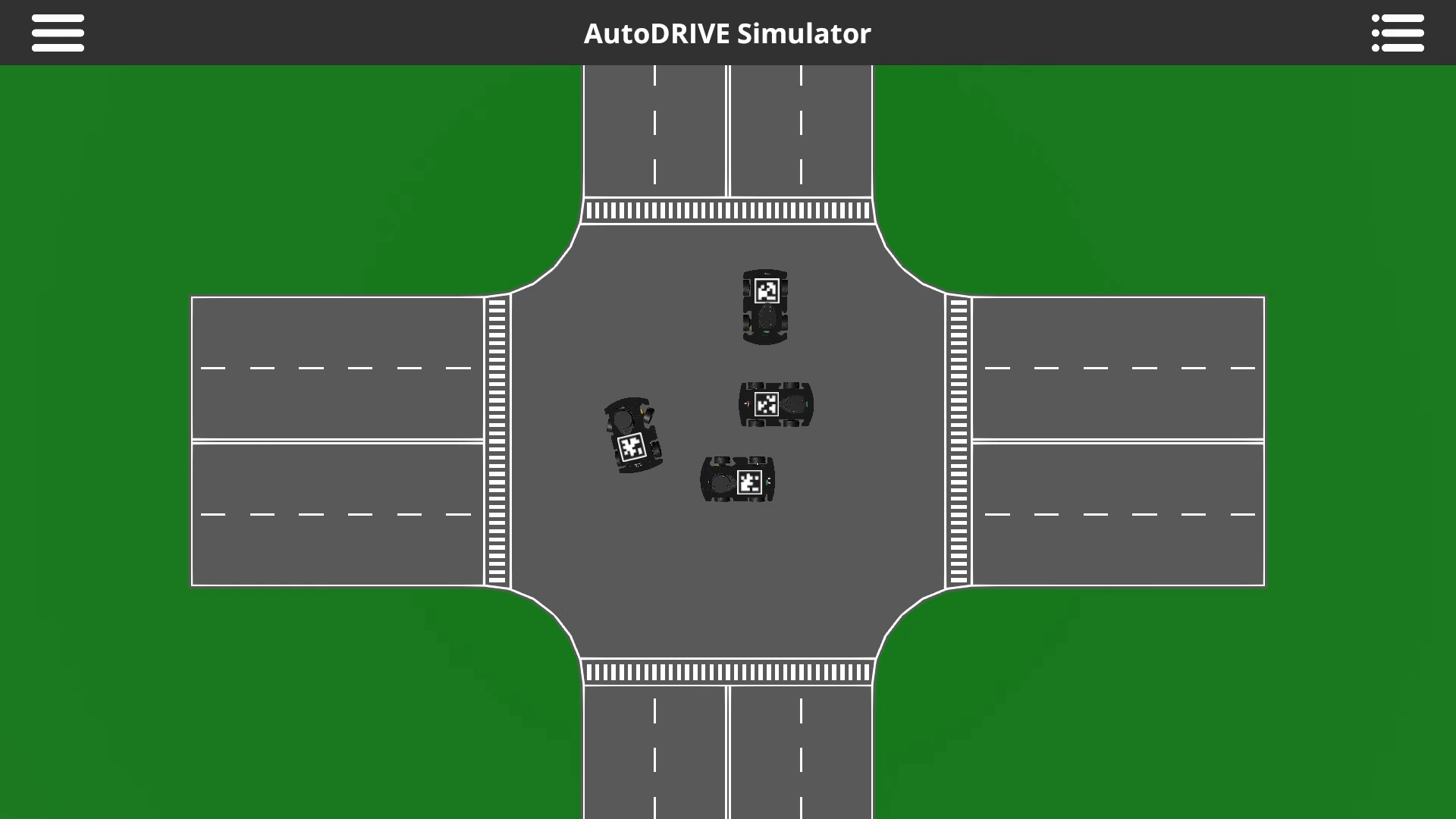}
			\caption{}
			\label{fig4b}
		\end{subfigure}
		\hfill
		\begin{subfigure}[b]{0.16\linewidth}
			\centering
			\includegraphics[width=\linewidth]{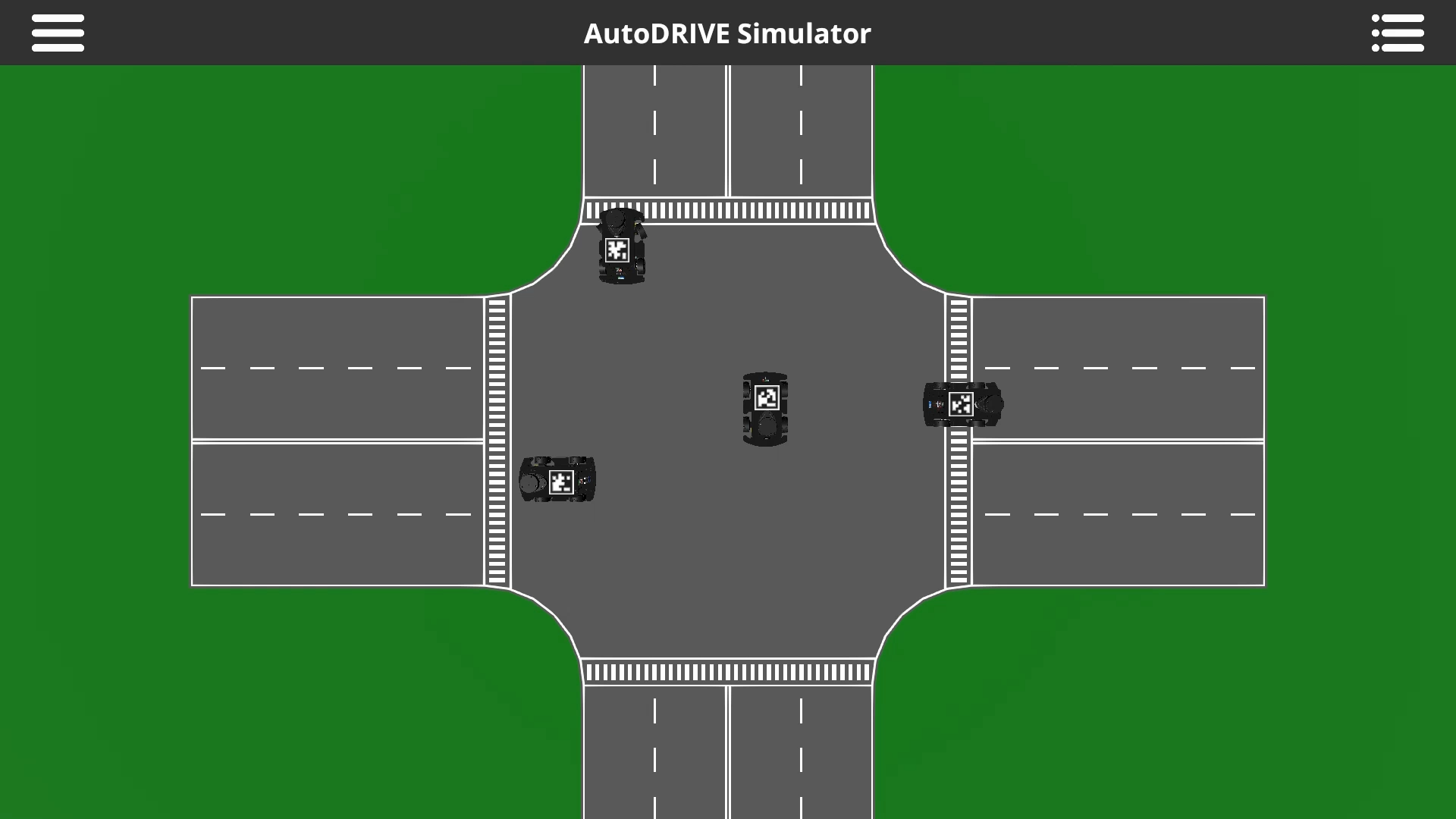}
			\caption{}
			\label{fig4c}
		\end{subfigure}
		\hfill
		\begin{subfigure}[b]{0.16\linewidth}
			\centering
			\includegraphics[width=\linewidth]{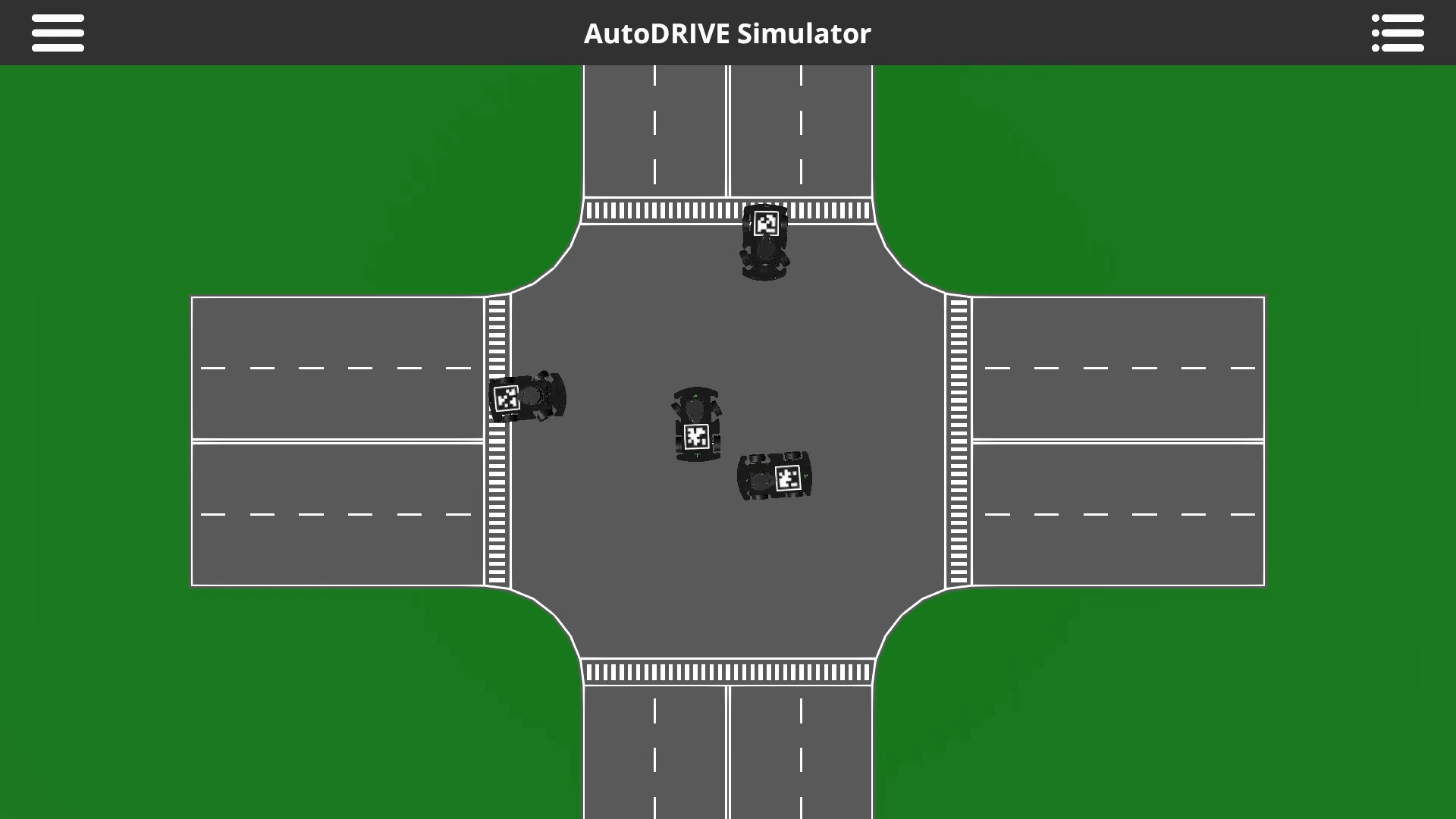}
			\caption{}
			\label{fig4d}
		\end{subfigure}
		\hfill
		\begin{subfigure}[b]{0.16\linewidth}
			\centering
			\includegraphics[width=\linewidth]{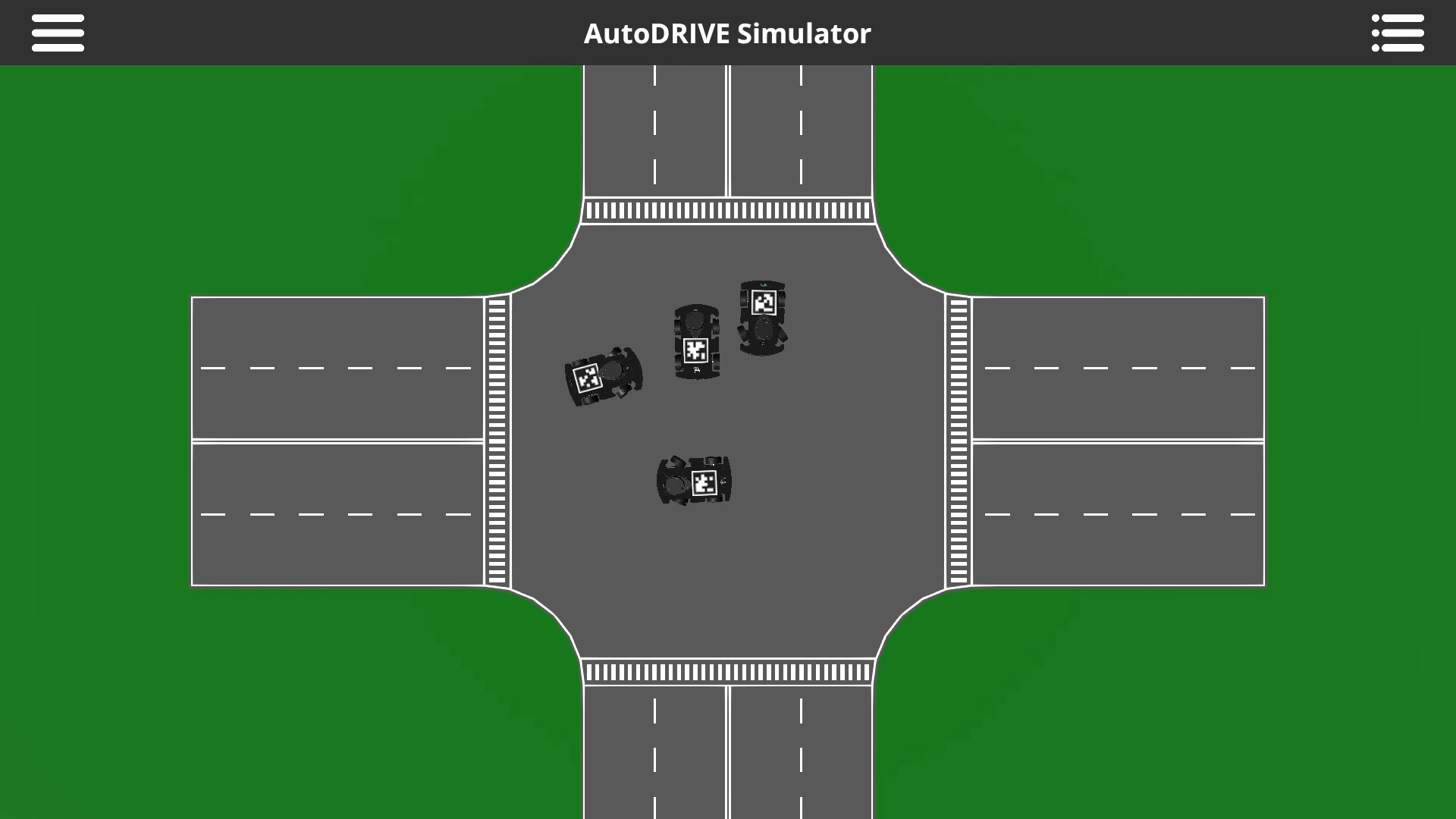}
			\caption{}
			\label{fig4e}
		\end{subfigure}
		\hfill
		\begin{subfigure}[b]{0.16\linewidth}
			\centering
			\includegraphics[width=\linewidth]{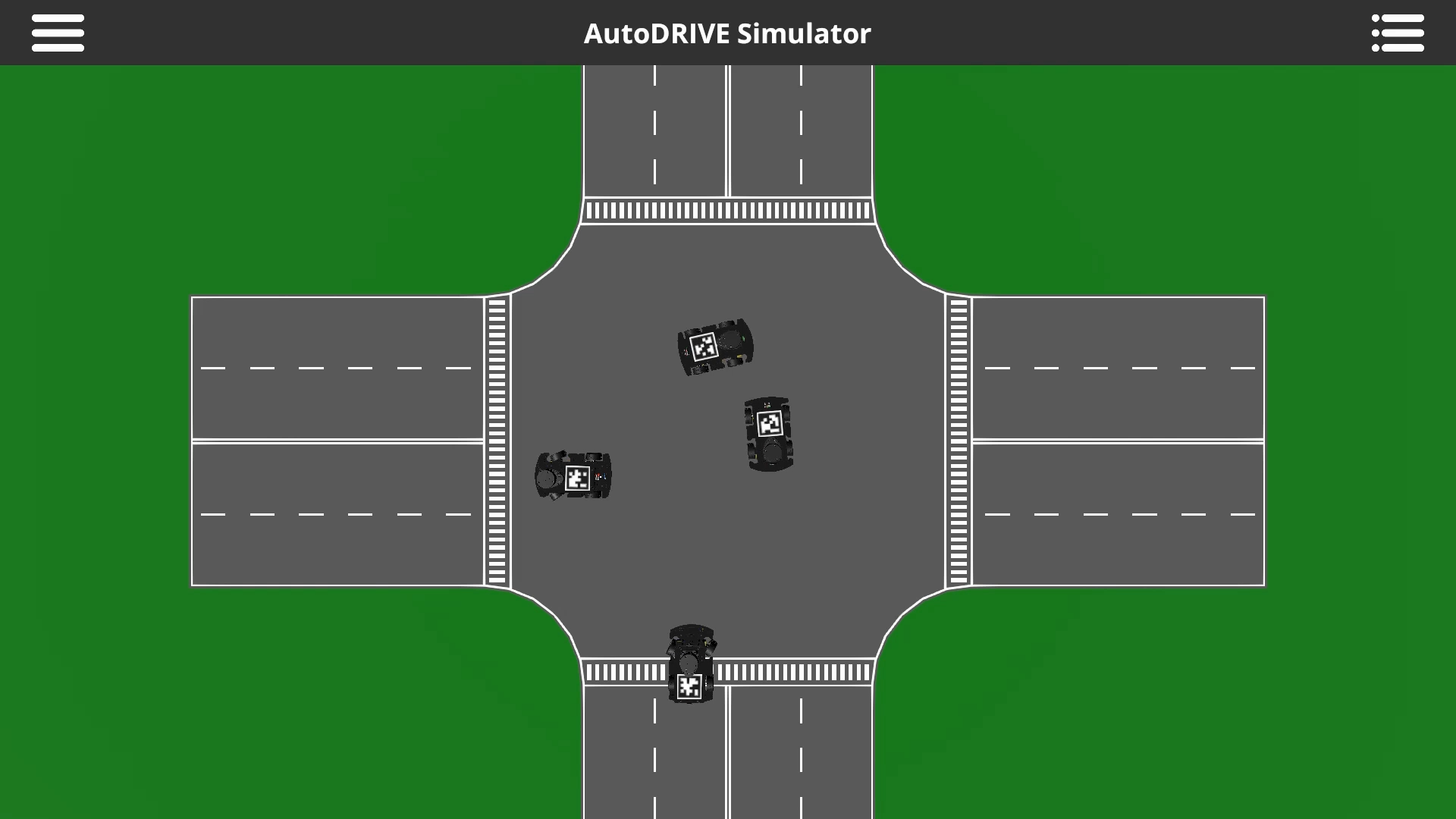}
			\caption{}
			\label{fig4f}
		\end{subfigure}
		\caption{Deployment results for (a)-(c) single-agent and (d)-(f) multi-agent intersection traversal scenarios: (a) and (d) denote first frozen snapshot, (b) and (e) denote second frozen snapshot, while (c) and (f) denote third frozen snapshot.}
		\label{fig4}
	\end{figure*}
	
	\subsection{Training}
	\label{Sub-Section: Training I}
	
	At each time step $t$, each parallelized agent $i$ collected an observation vector $o_t^i$ and an extrinsic reward $r_t^i$. Based on these inputs, it took an action $a_t^i$ determined by the policy $\pi_\theta$, which was continually updated to maximize the expected future discounted reward (refer Fig. \ref{fig3}).
	
	This use-case employed a fully connected neural network (FCNN) as a function approximator for $\pi_\theta \left ( a_t | o_t \right )$. The network had $\mathbb{R}^{14}$ inputs, $\mathbb{R}^{1}$ outputs, and three hidden layers with 128 neural units each. The policy parameters $\theta \in \mathbb{R}^d$ were defined in terms of the network's parameters. The policy was trained to predict steering commands directly based on collected observations, utilizing the proximal policy optimization (PPO) algorithm \cite{PPO2017}.
	
	\subsection{Deployment}
	\label{Sub-Section: Deployment I}
	
	The trained policies were deployed onto the simulated vehicles, separately for both single-agent and multi-agent scenarios. As previously mentioned, the single-agent scenario was relatively deterministic, and the ego vehicle could safely traverse the intersection in most cases. In contrast, the multi-agent scenario was highly stochastic, resulting in a significantly lower success rate, especially with all vehicles navigating the intersection safely simultaneously.
	
	Fig.\hyperref[fig4]{\ref*{fig4}(a)-(c)} present three key stages of the single-agent intersection traversal scenario. The first stage depicts the ego vehicle approaching the conflict zone, where it could potentially collide with peer vehicles. The second stage shows the vehicle executing a left turn to avoid collisions. Finally, the third stage illustrates the vehicle performing a subtle right turn to reach its goal. Fig.\hyperref[fig4]{\ref*{fig4}(d)-(f)} display three critical stages of the multi-agent intersection traversal scenario. In the first frame, vehicles 1 and 4 successfully avoid collision. The second frame showcases vehicle 1 finding a gap between vehicles 2 and 3 to reach its goal. In the third frame, vehicles 2 and 3 evade collision, while vehicle 4 approaches its goal, and vehicle 1 is re-spawned.
	
	% %%%%%%%%%%%%%%%%%%%%%%%%%%%%%%%%%%%%%%%%%%%%%%%%%%%%%%%%%%%%%%%%%%%%%%%%%%%%%%%%
	
	\section{Competitive Multi-Agent Scenario}
	\label{Section: Competitive Multi-Agent Scenario}
	
	Inspired by \cite{AutoRACE-2021}, this use-case encompassed a multi-agent learning scenario, where each agent's objective was minimizing its lap time without colliding with the track or its opponent. Each vehicle possessed intrinsic state information and sparse LIDAR measurements; no state information was shared among the competing vehicles. The entire use-case was developed using the same integrated ML framework mentioned in Section \ref{Section: Cooperative Multi-Agent Scenario}.
	
	\subsection{Problem Formulation}
	\label{Sub-Section: Problem Formulation II}
	
	This case-study addressed the problem of autonomous racing in an unknown environment. The exact structure/map of the environment was not known to any agent. Consequently, this decision-making problem was also framed as a POMDP, which captured hidden state information through environmental observations.
	
	We adopted an equally-weighted hybrid imitation-reinforcement learning architecture to progressively inculcate autonomous driving and racing behaviors into the agents. Consequently, we recorded 5 laps worth of independent demonstration datasets for each agent by manually driving the vehicles in sub-optimal trajectories within a single-agent setting. We hypothesised that such a hybrid learning architecture would guide the agents' exploration, thereby reducing training time significantly.
	
	\begin{figure*}[t]
		\centering
		\begin{subfigure}[b]{0.16\linewidth}
			\centering
			\includegraphics[width=\linewidth]{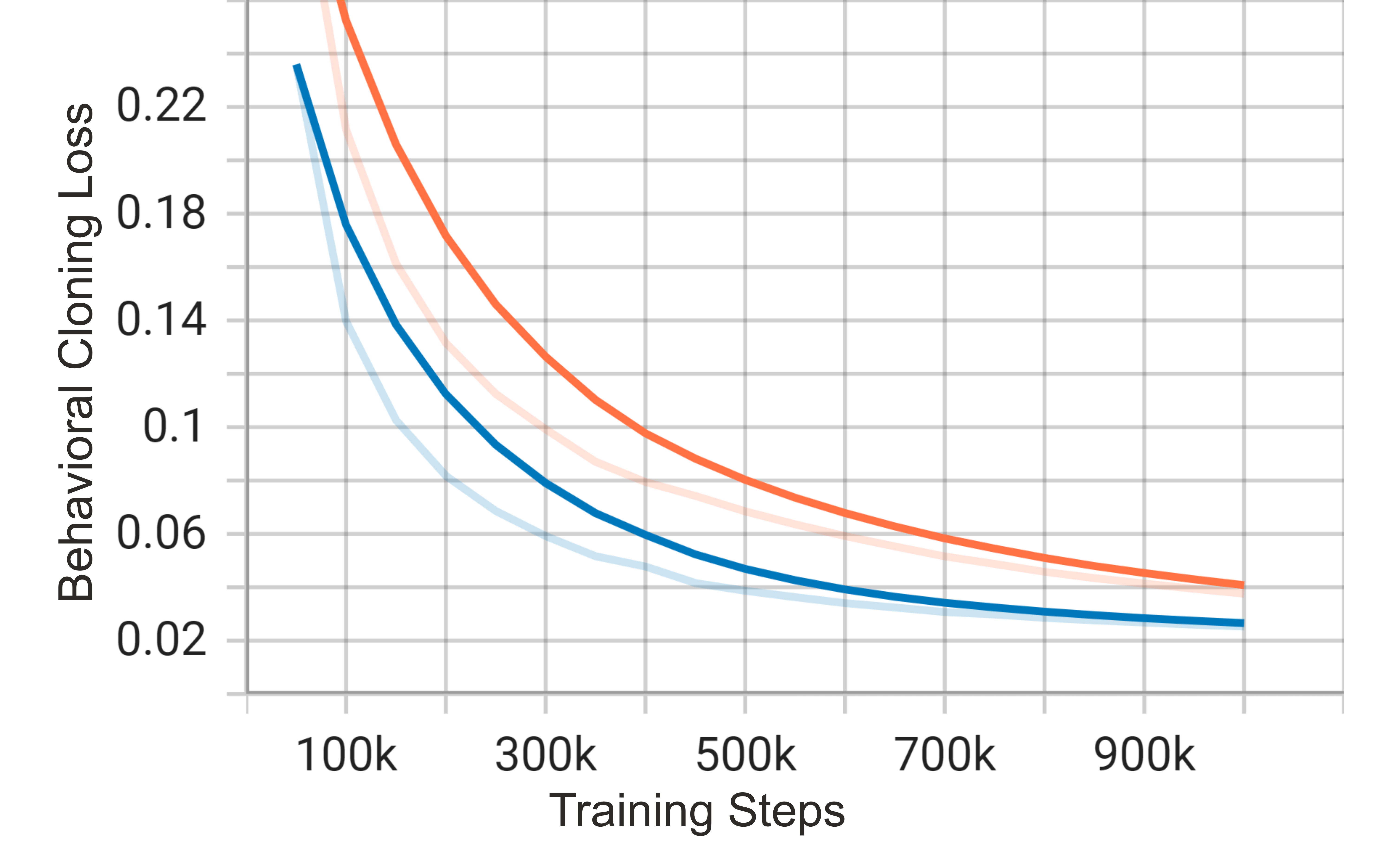}
			\caption{}
			\label{fig5a}
		\end{subfigure}
		\hfill
		\begin{subfigure}[b]{0.16\linewidth}
			\centering
			\includegraphics[width=\linewidth]{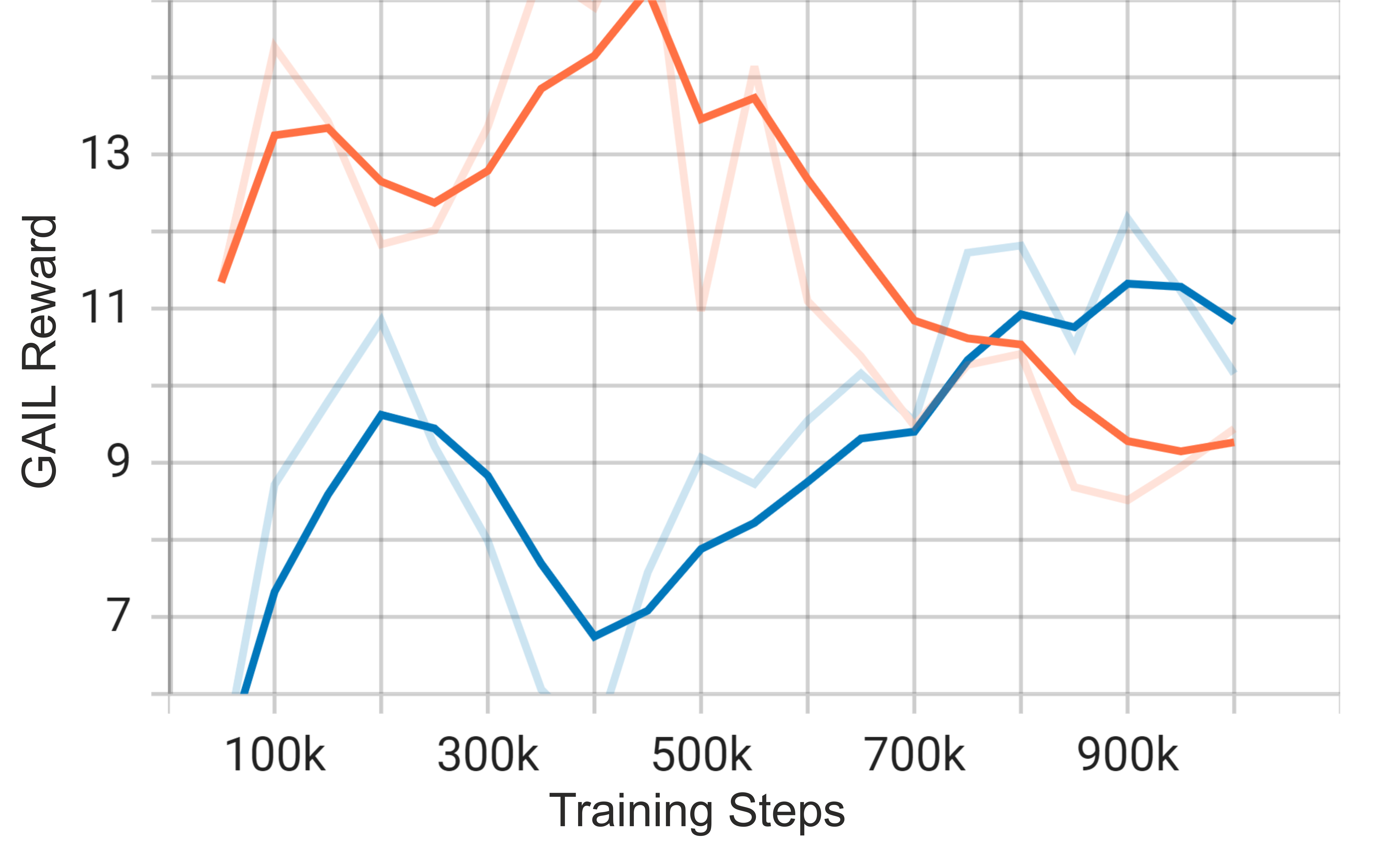}
			\caption{}
			\label{fig5b}
		\end{subfigure}
		\hfill
		\begin{subfigure}[b]{0.16\linewidth}
			\centering
			\includegraphics[width=\linewidth]{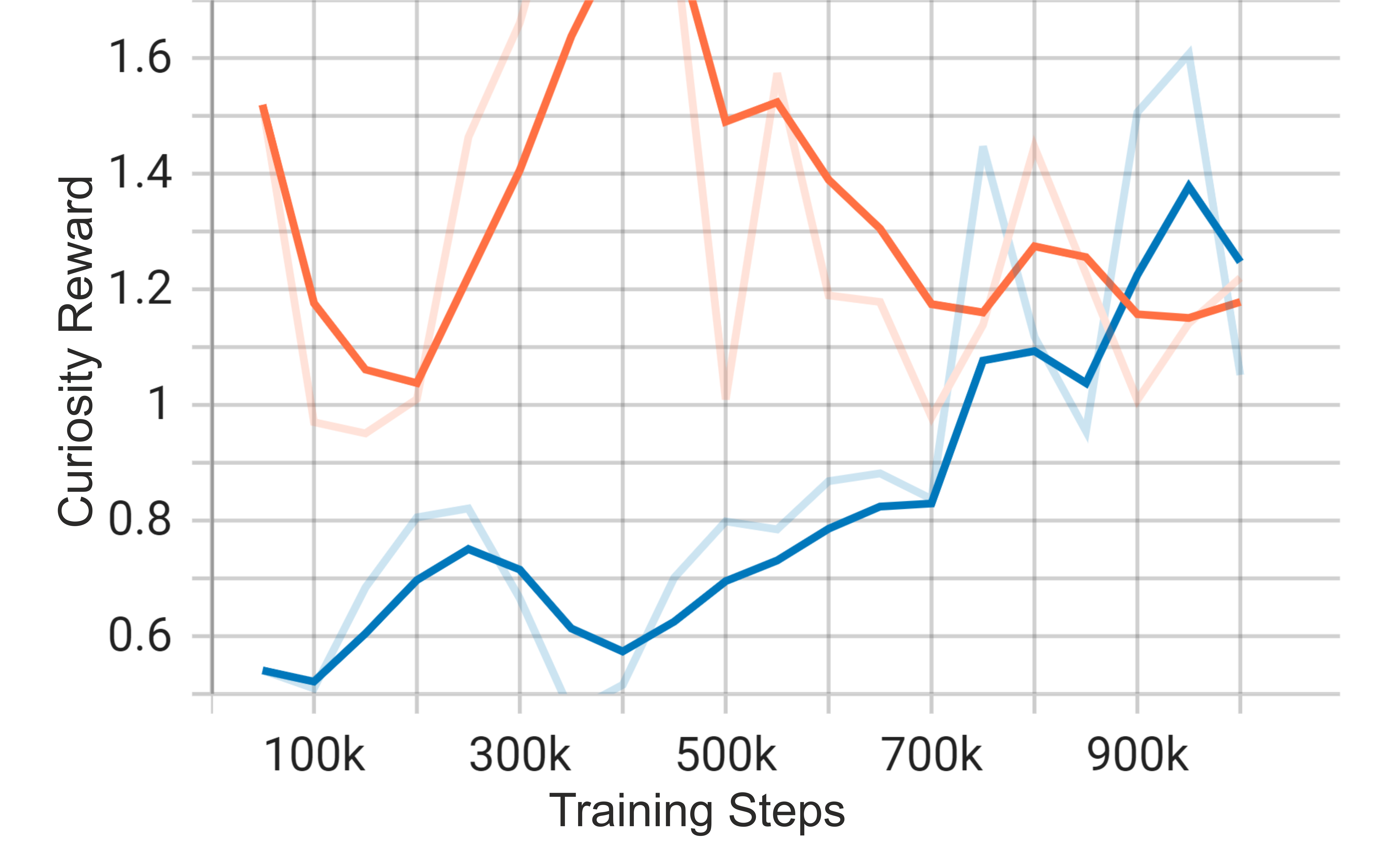}
			\caption{}
			\label{fig5c}
		\end{subfigure}
		\hfill
		\begin{subfigure}[b]{0.16\linewidth}
			\centering
			\includegraphics[width=\linewidth]{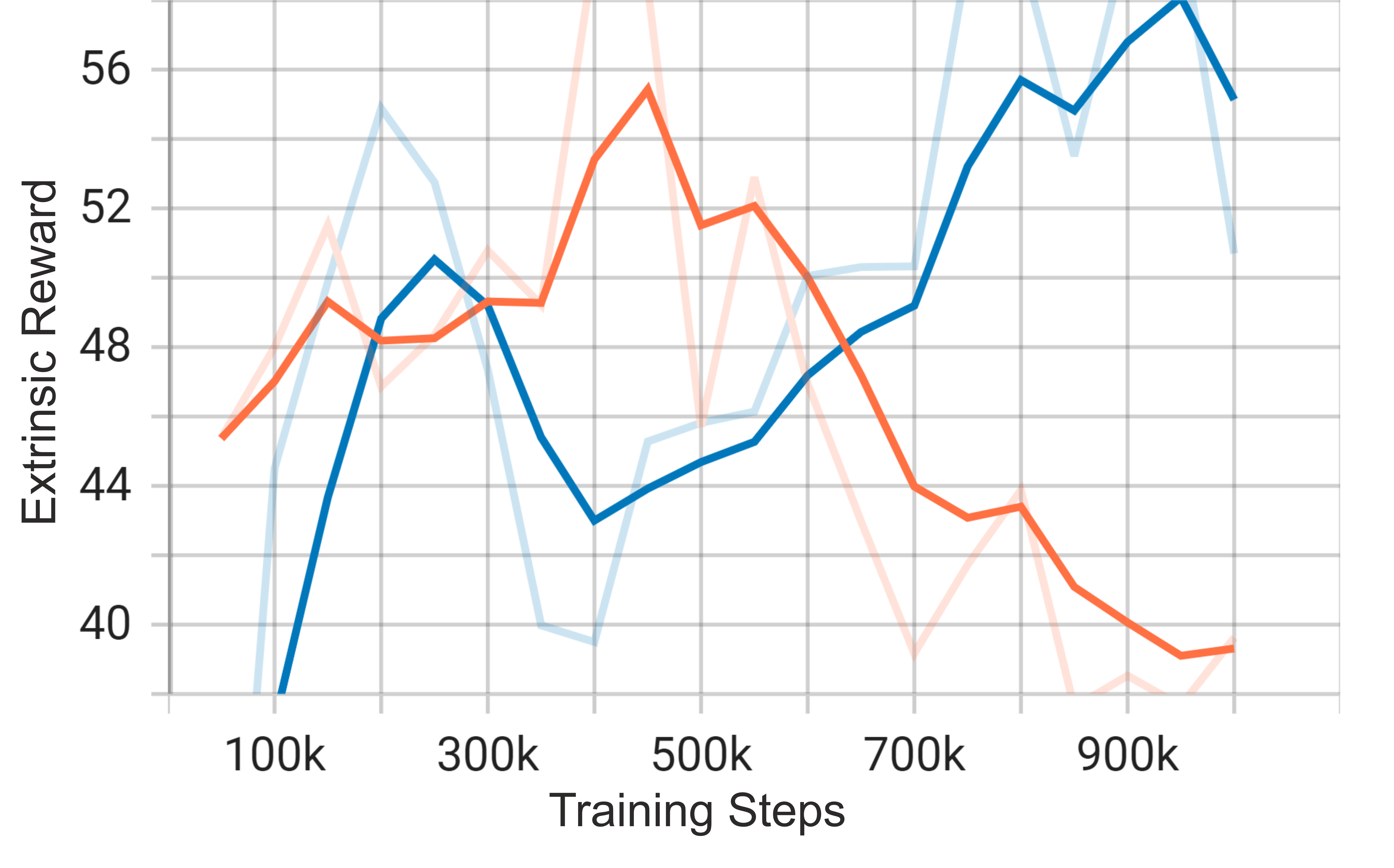}
			\caption{}
			\label{fig5d}
		\end{subfigure}
		\hfill
		\begin{subfigure}[b]{0.16\linewidth}
			\centering
			\includegraphics[width=\linewidth]{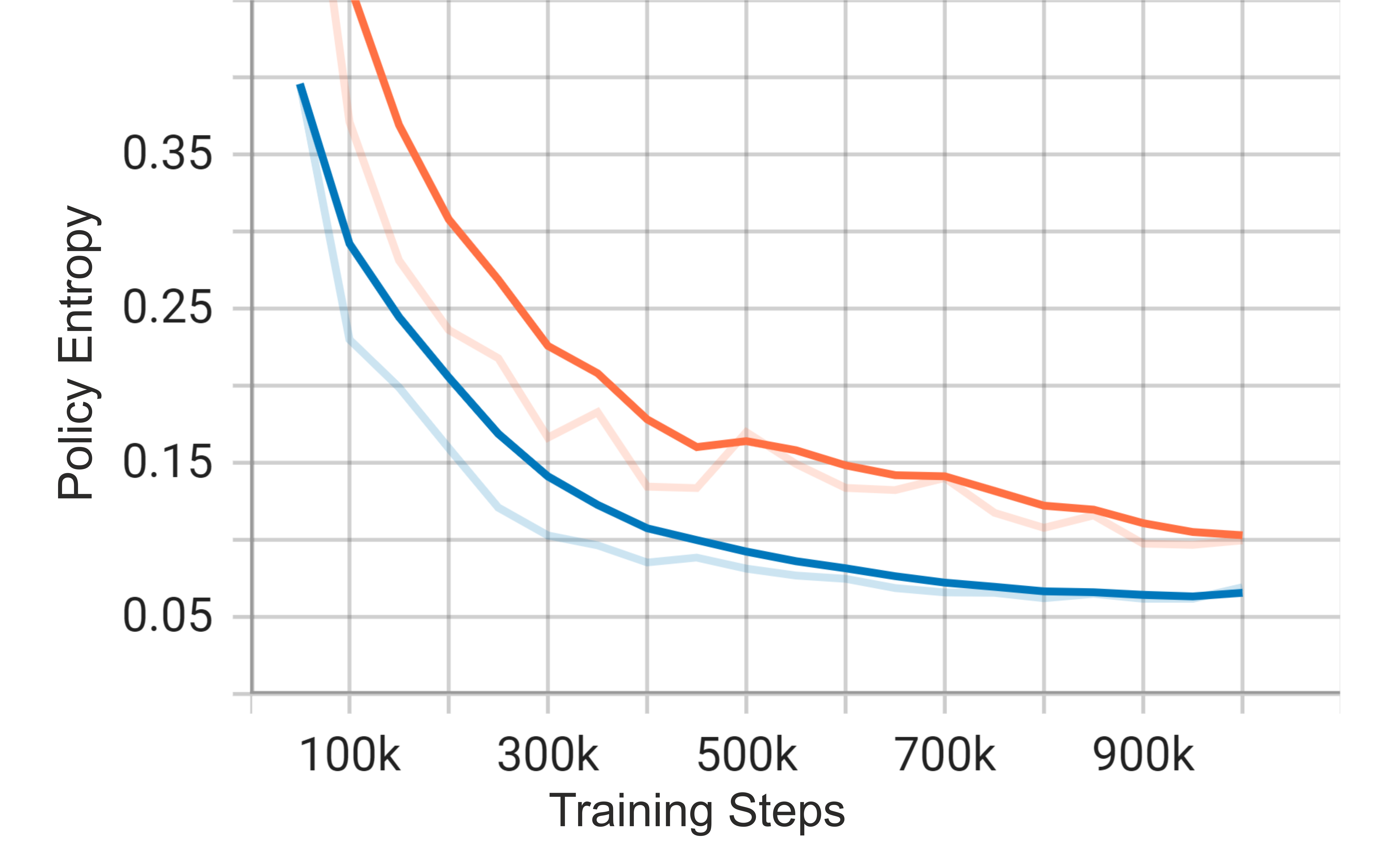}
			\caption{}
			\label{fig5e}
		\end{subfigure}
		\hfill
		\begin{subfigure}[b]{0.16\linewidth}
			\centering
			\includegraphics[width=\linewidth]{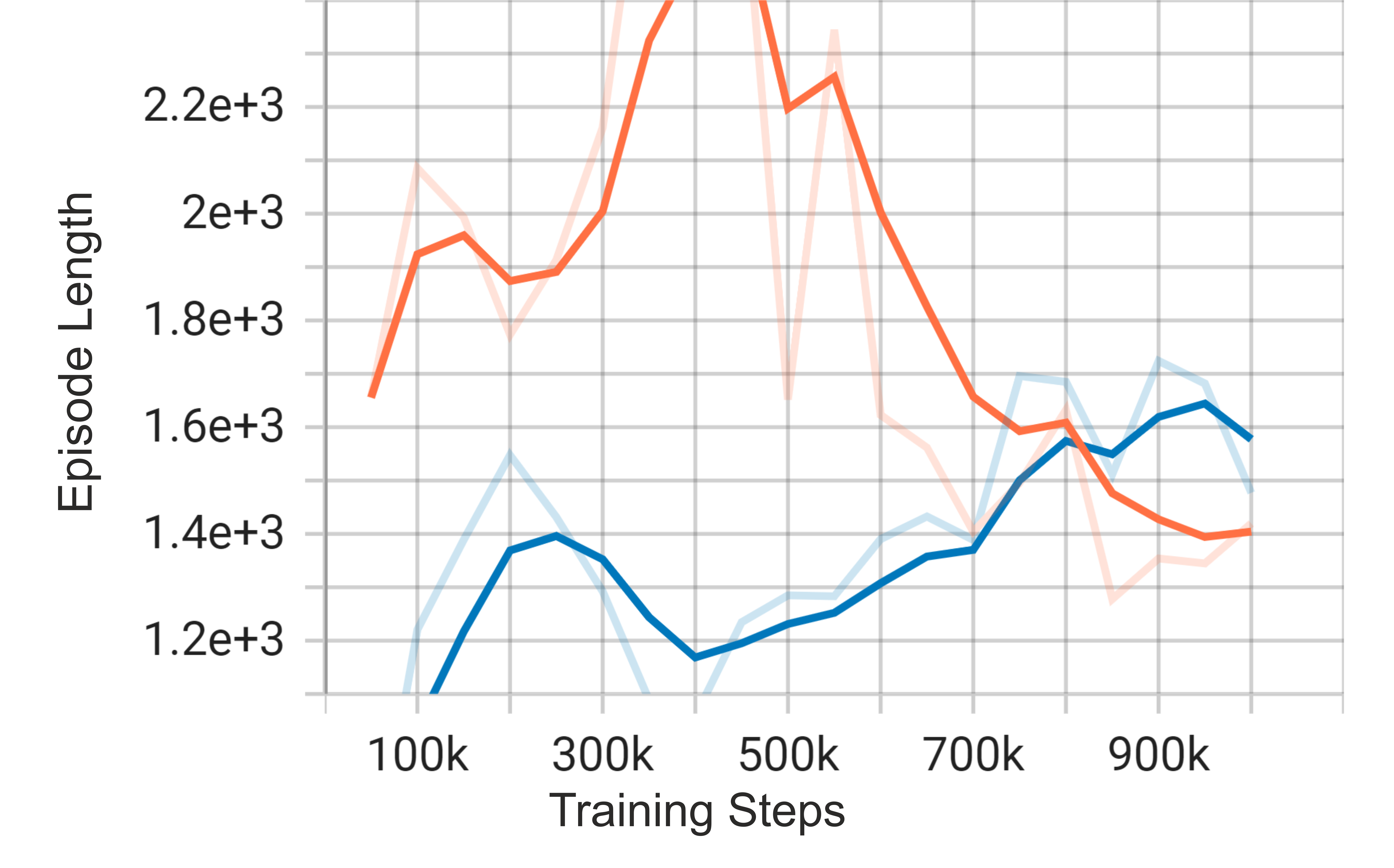}
			\caption{}
			\label{fig5f}
		\end{subfigure}
		\caption{Training results for multi-agent autonomous racing: (a) denotes BC loss, (b) denotes GAIL reward, (c) denotes curiosity reward, (d) denotes extrinsic reward, (e) denotes policy entropy, and (f) denotes episode length w.r.t. training steps.}
		\label{fig5}
	\end{figure*}
	
	\begin{figure*}[t]
		\centering
		\begin{subfigure}[b]{0.16\linewidth}
			\centering
			\includegraphics[width=\linewidth]{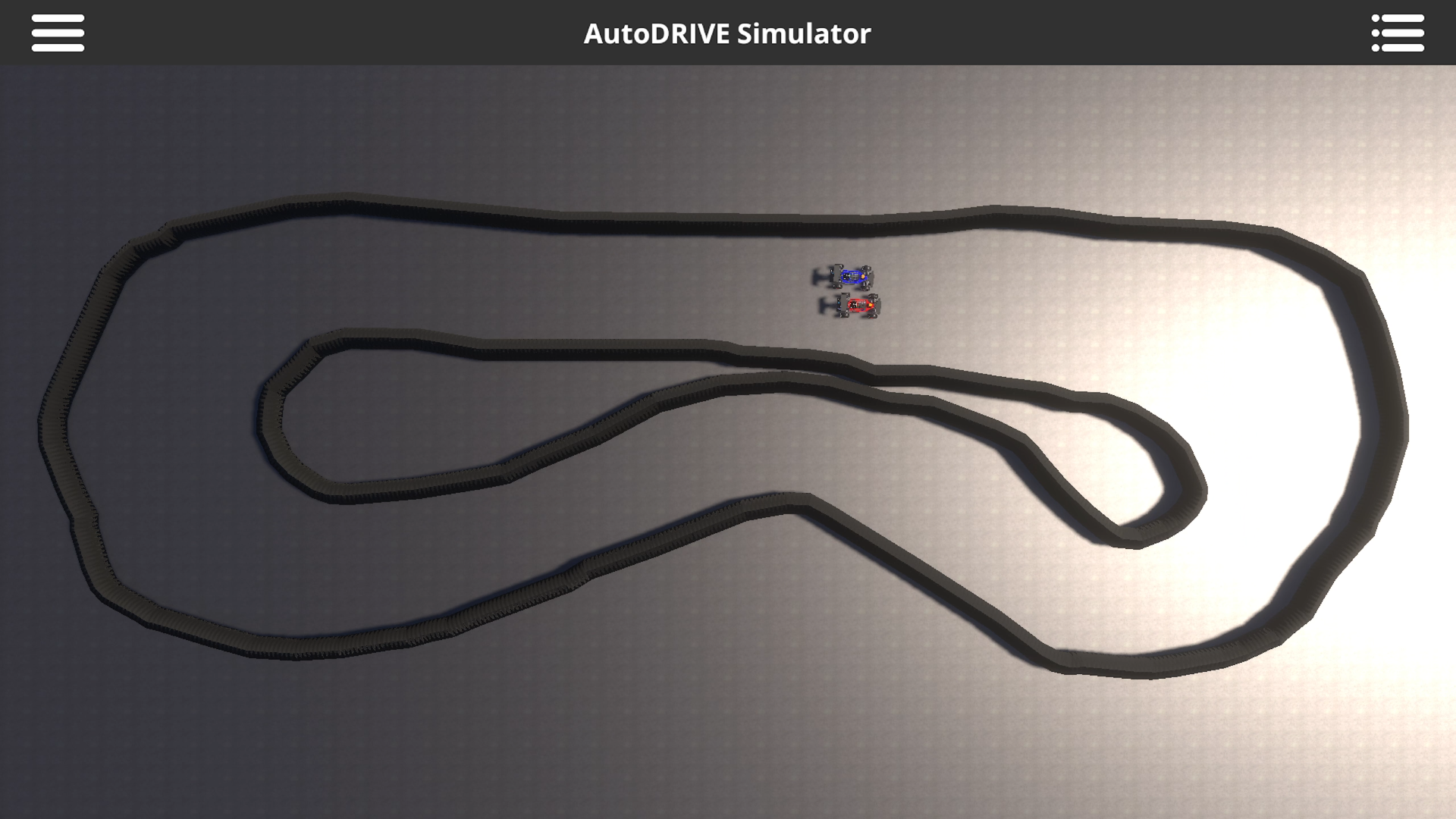}
			\caption{}
			\label{fig6a}
		\end{subfigure}
		\hfill
		\begin{subfigure}[b]{0.16\linewidth}
			\centering
			\includegraphics[width=\linewidth]{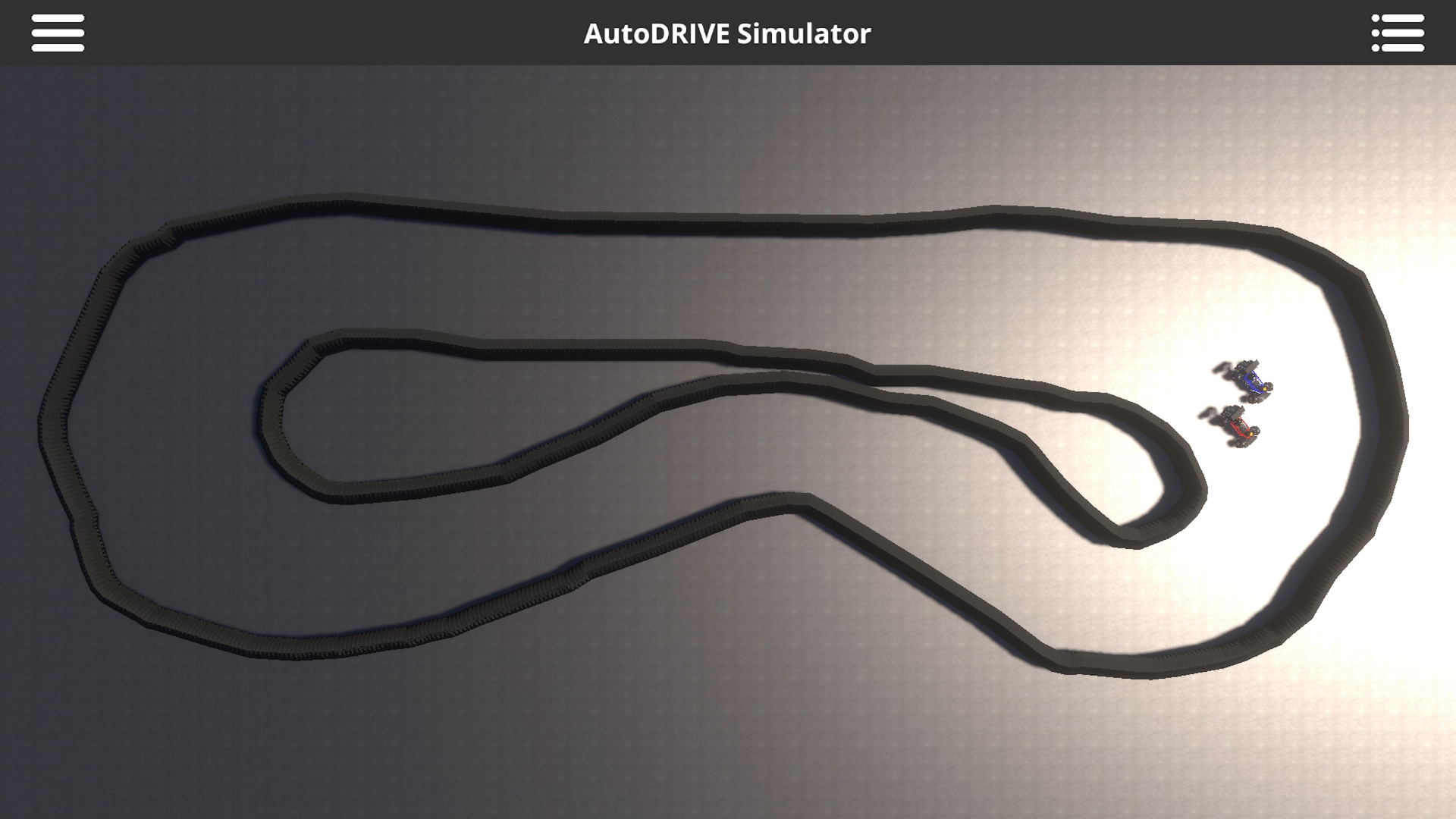}
			\caption{}
			\label{fig6b}
		\end{subfigure}
		\hfill
		\begin{subfigure}[b]{0.16\linewidth}
			\centering
			\includegraphics[width=\linewidth]{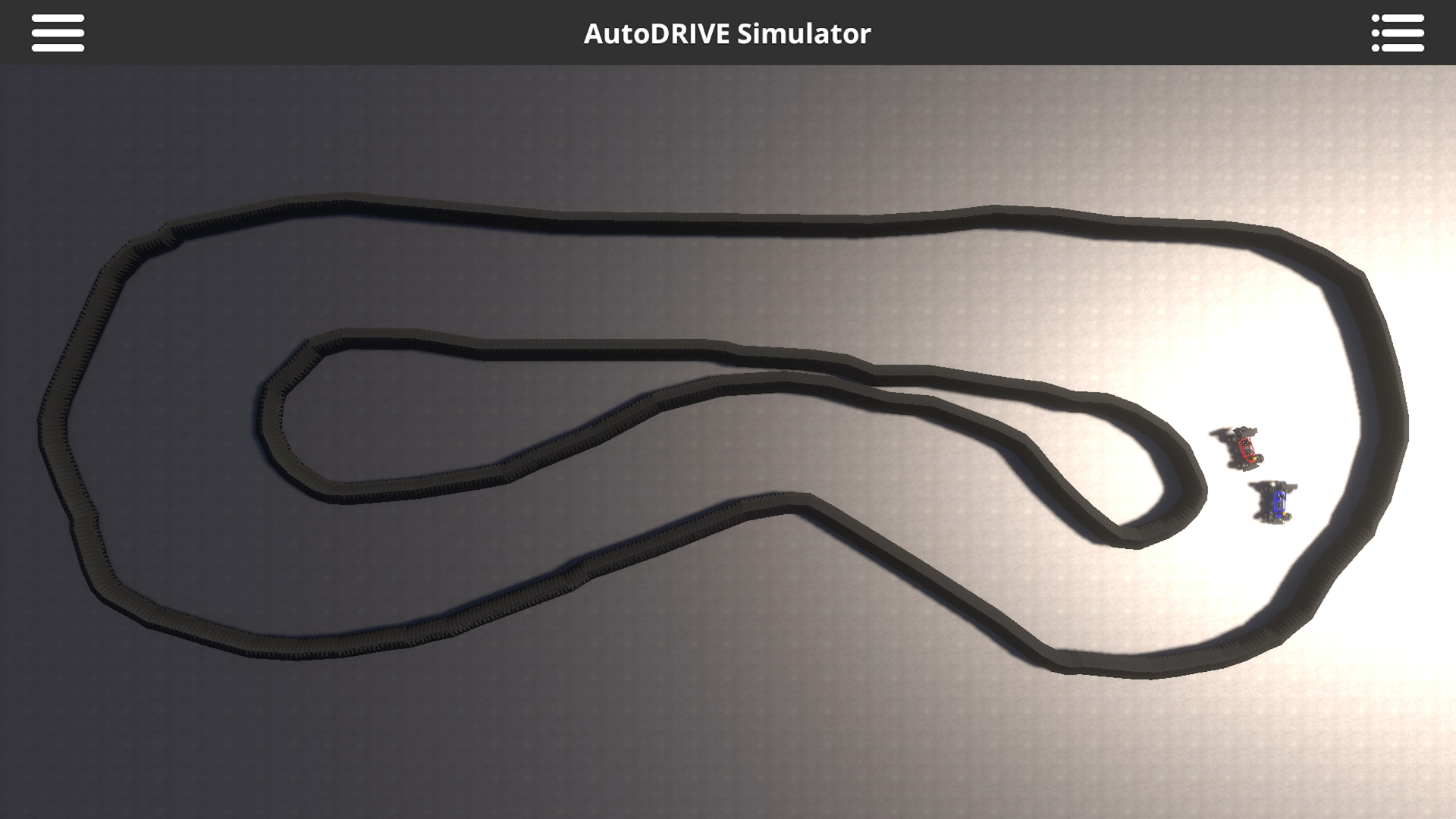}
			\caption{}
			\label{fig6c}
		\end{subfigure}
		\hfill
		\begin{subfigure}[b]{0.16\linewidth}
			\centering
			\includegraphics[width=\linewidth]{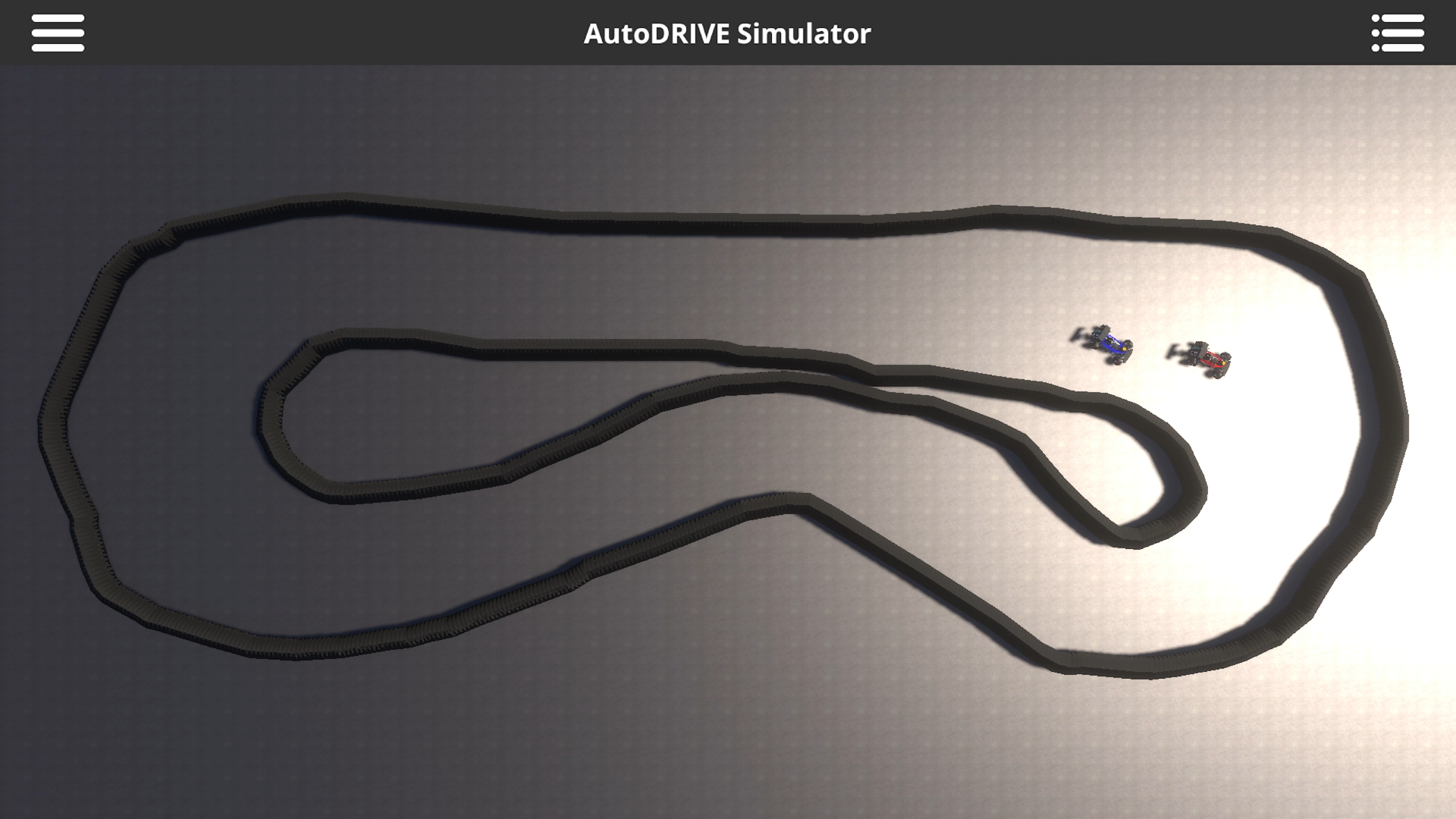}
			\caption{}
			\label{fig6d}
		\end{subfigure}
		\hfill
		\begin{subfigure}[b]{0.16\linewidth}
			\centering
			\includegraphics[width=\linewidth]{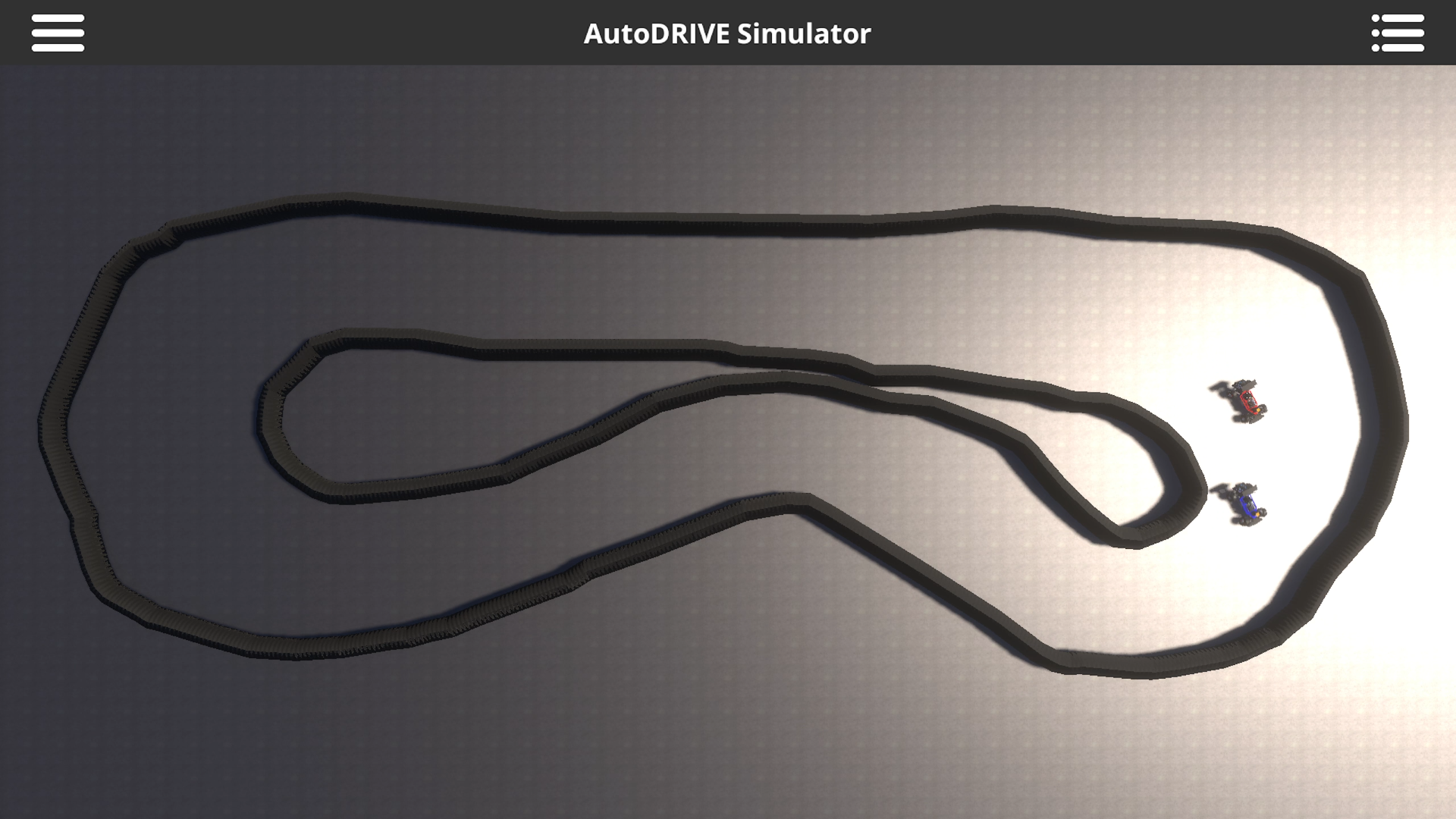}
			\caption{}
			\label{fig6e}
		\end{subfigure}
		\hfill
		\begin{subfigure}[b]{0.16\linewidth}
			\centering
			\includegraphics[width=\linewidth]{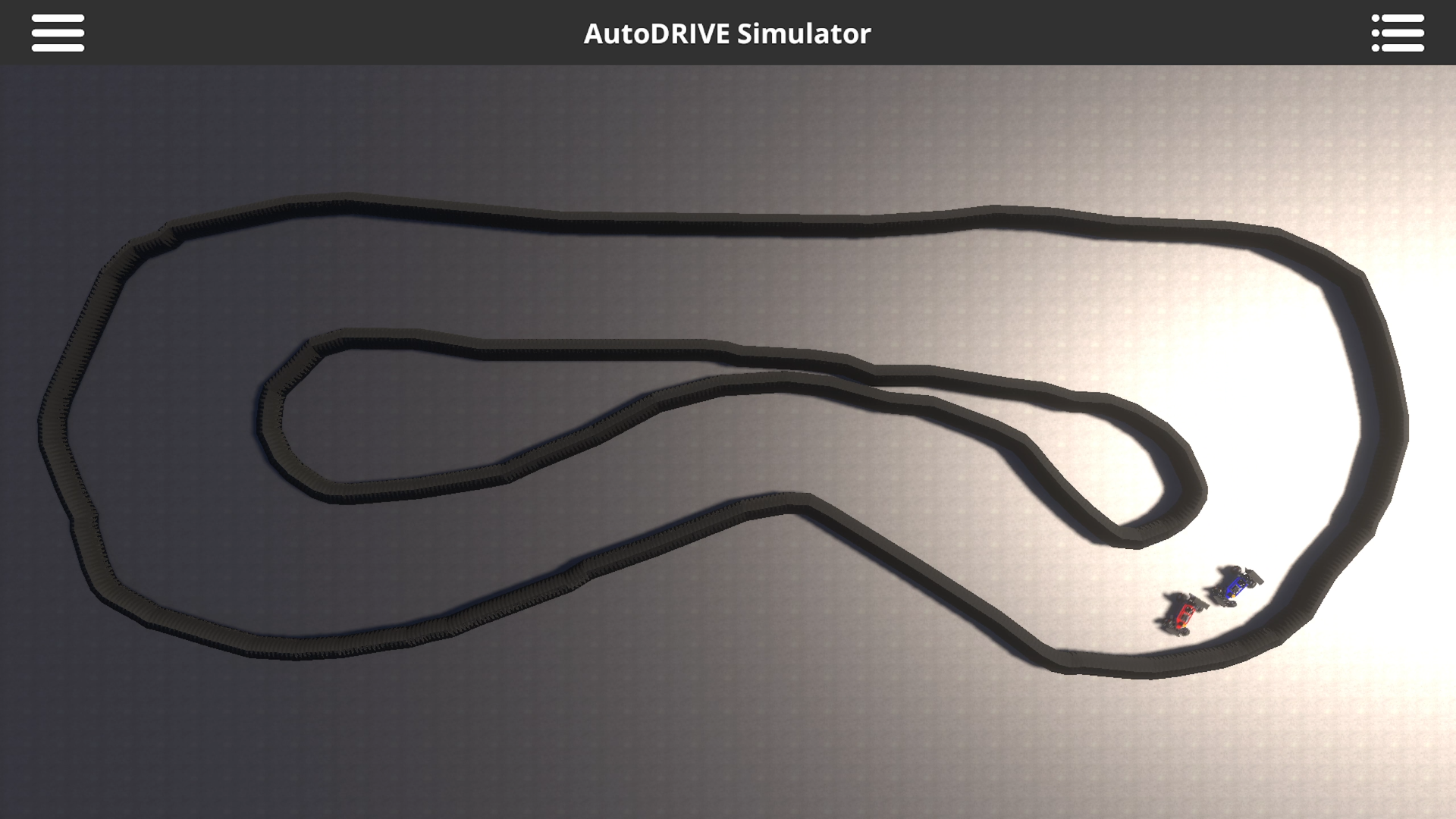}
			\caption{}
			\label{fig6f}
		\end{subfigure}
		\caption{Deployment results for multi-agent autonomous racing: (a)-(c) denote three frozen snapshots of a block-block-overtake sequence, and (d)-(f) denote three frozen snapshots of a let-pass-and-overtake sequence.}
		\label{fig6}
	\end{figure*}
	
	\subsection{Observation Space}
	\label{Sub-Section: Observation Space II}
	
	At each time step $t$, the agent collected a vectorized observation, as shown in Equation \ref{eqn1}, from the environment. These observations were obtained using velocity estimation and exteroceptive ranging modalities mounted on the virtual vehicle(s):
	
	\begin{equation}
		o_t^i = \left [ v_t^i, m_t^i \right ] \in \mathbb{R}^{28}
		\label{eqn1}
	\end{equation}
	
	Here, $v_t^i \in \mathbb{R}^{1}$ represents the forward velocity of $i$-th agent, and $m_t^i = \left [ ^{1}m_t^i, ^{2}m_t^i, \cdots, ^{27}m_t^i \right ] \in \mathbb{R}^{27}$ is the measurement vector providing 27 range readings up to 10 meters. These readings are uniformly distributed over 270$^{\circ}$ around each side of the heading vector, spaced 10$^{\circ}$ apart. These observations were then input into a deep neural network policy denoted as $\pi_{\theta}$, where $\theta \in \mathbb{R}^d$ denotes the policy parameters.
	
	\subsection{Action Space}
	\label{Sub-Section: Action Space II}
	
	The policy mapped the observations $o_t$ directly to an appropriate action $a_t$, as expressed in Equation \ref{eqn2}:
	
	\begin{equation}
		a_t^i = \left [ \tau_t^i, \delta_t^i \right ] \in \mathbb{R}^{2}
		\label{eqn2}
	\end{equation}
	
	Here, $\tau_t^i \in \left \{ 0.1, 0.5, 1.0 \right \}$ represent the discrete throttle commands at 10\%, 50\% and 100\% PWM duty cycles for torque limited (85.6 N-m) drive actuators, and $\delta_t^i \in \left \{ -1, 0, 1 \right \}$ represent the discrete steering commands for left, straight, and right turns, respectively.
	
	\subsection{Reward Function}
	\label{Sub-Section: Reward Function II}
	
	The policy $\pi_{\theta}$ was optimized based on Behavioral Cloning (BC) \cite{bain1995}, Generative Adversarial Imitation Learning (GAIL) $^{g}r_t$ reward \cite{ho2016}, curiosity $^{c}r_t$ reward \cite{pathak2017} as well as an extrinsic reward $^{e}r_t$ (as detailed in Equation \ref{eqn3}).
	
	Particularly, The agent received a reward of $r_{checkpoint}=+0.01$ for passing each of the 19 checkpoints $c_i$, where $i \in \left [ \text{A}, \text{B}, \cdots, \text{S} \right ]$ on the racetrack, $r_{lap}=+0.1$ upon completing a lap, $r_{best\:lap}=+0.7$ upon achieving a new best lap time, and a penalty of $r_{collision}=-1$ for colliding with any of the track walls $w_j$, where $j \in \mathbb{R}^{n}$. Additionally, the agent received continuous rewards proportional to its velocity $v_t$, encouraging it to optimize its trajectory spatio-temporally.
	
	\begin{equation}
		^{e}r_t^i  =  
		\begin{cases}
			r_{collision} & \text{if collision occurs} \\
			r_{checkpoint} & \text{if checkpoint is passed} \\
			r_{lap} & \text{if completed lap} \\
			r_{best\:lap} & \text{if new best lap time is achieved} \\
			0.01*v_t^i & \text{otherwise}
		\end{cases}
		\label{eqn3}
	\end{equation}
	
	\subsection{Training}
	\label{Sub-Section: Training II}
	
	The policy $\pi_\theta$ was optimized to maximize the expected future discounted reward (GAIL, curiosity and extrinsic rewards), while also minimizing the BC loss (refer Fig. \ref{fig5}).
	
	This use-case also employed a fully connected neural network (FCNN) as a function approximator for $\pi_\theta \left ( a_t | o_t \right )$. The network had $\mathbb{R}^{28}$ inputs, $\mathbb{R}^{2}$ outputs, and three hidden layers with 128 neural units each. The policy parameters $\theta \in \mathbb{R}^d$ were defined in terms of the network's parameters. The policy was trained to predict throttle and steering commands directly based on collected observations, utilizing the proximal policy optimization (PPO) algorithm \cite{PPO2017}.
	
	\subsection{Deployment}
	\label{Sub-Section: Deployment II}
	
	The trained policies were deployed onto the respective simulated vehicles, which were made to race head-to-head on the same track with a phase-shifted initialization (as in real F1TENTH competitions).
	
	Fig.\hyperref[fig6]{\ref*{fig6}(a)-(c)} present three snapshots of a block-block-overtake sequence, wherein the red agent kept blocking the blue agent throughout the straight, but the blue agent took a wider turn with higher velocity and took advantage of its under-steer characteristic to cut in front of the red agent and overtake it. Fig.\hyperref[fig6]{\ref*{fig6}(d)-(f)} display three snapshots of a let-pass-and-overtake sequence, wherein the blue agent found a gap between the red agent and inside edge of the track and opportunistically overtook it. However, due to its under-steering characteristic, it went wider in the corner, thereby giving the red agent an opportunity to overtake it and re-claim the leading position.
	
	\section{Conclusion}
	\label{Section: Conclusion}
	
	This work presented a multi-agent reinforcement learning framework for imbibing cooperative and competitive behaviors within autonomous vehicles using the real2sim approach. We discussed representative case-studies for each behavior type and analyzed the training and deployment results. A natural extension of this work would be to analyze the sim2real \cite{samak2023sim2real} transfer of these trained policies.
	
	%%%%%%%%%%%%%%%%%%%%%%%%%%%%%%%%%%%%%%%%%%%%%%%%%%%%%%%%%%%%%%%%%%%%%%%%%%%%%%%%
	%\addtolength{\textheight}{-12cm}  % This command serves to balance the column lengths
	
	%%%%%%%%%%%%%%%%%%%%%%%%%%%%%%%%%%%%%%%%%%%%%%%%%%%%%%%%%%%%%%%%%%%%%%%%%%%%%%%%
	
	\bibliographystyle{IEEEtran}
	\bibliography{References}
	
\end{document}